\documentclass{article}

    \usepackage[preprint]{neurips_2023}
\usepackage{amsthm}
\usepackage{natbib}
\setcitestyle{numbers,square}
\usepackage[utf8]{inputenc} % allow utf-8 input
\usepackage[T1]{fontenc}    % use 8-bit T1 fonts
\usepackage{hyperref}       % hyperlinks
\usepackage{url}            % simple URL typesetting
\usepackage{booktabs}       % professional-quality tables
\usepackage{amsfonts}       % blackboard math symbols
\usepackage{nicefrac}       % compact symbols for 1/2, etc.
\usepackage{microtype}      % microtypography
\usepackage{xcolor}         % colors
\usepackage{multicol}
\usepackage{enumerate}
\usepackage[shortlabels]{enumitem}
\usepackage{mdframed}
\usepackage{colortbl}
\usepackage[most]{tcolorbox}   
\usepackage{aligned-overset}
\usepackage{bbm}
\usepackage{dsfont}
\usepackage{enumitem}
\usepackage{wrapfig}
\usepackage{colortbl,booktabs}
\usepackage{xcolor}
\usepackage{caption}
\usepackage{subcaption}
\usepackage{wrapfig}

\usepackage{tabularray}

\usepackage{chemformula}

\usepackage[ruled,linesnumbered]{algorithm2e}
\usepackage[capitalize]{cleveref}
\crefname{section}{Sec.}{Secs.}
\Crefname{section}{Section}{Sections}
\Crefname{table}{Table}{Tables}
\crefname{table}{Tab.}{Tabs.}

\newtheorem{theorem}{Theorem}

\newtheorem{lemma}{Lemma}

\newtheorem{definition}{Definition}
\newtheorem{assumption}{Assumption}
\newtheorem{remark}{Remark}

\usepackage{multirow}

\newcommand{\E}{\mathbb{E}}

\title{Towards More Suitable Personalization in Federated Learning via Decentralized Partial Model Training}

\author{
Yifan Shi\textsuperscript{\rm 1,†} 
\quad
Yingqi Liu\textsuperscript{\rm 2,†}
\quad
Yan Sun\textsuperscript{\rm 3}
\quad
Zihao Lin\textsuperscript{\rm 4}
\and
\textbf{Li Shen}\textsuperscript{\rm 5,}\thanks{Corresponding authors: Li Shen and Xueqian Wang. \quad†Equal contribution.}
\quad
\textbf{Xueqian Wang}\textsuperscript{\rm 1,*}
\quad
\textbf{Dacheng Tao}\textsuperscript{\rm 3}
\\
\textsuperscript{\rm 1}Tsinghua University; 
\textsuperscript{\rm 2}Nanjing University of Science and Technology\\
\textsuperscript{\rm 3}The University of Sydney; 
\textsuperscript{\rm 4}Virginia Tech; 
\textsuperscript{\rm 5}JD Explore Academy\\
{\tt\small shiyf21@mails.tsinghua.edu.cn;
lyq@njust.edu.cn; ysun9899@uni.sydney.edu.au 
}\\
{\tt\small zihaol@vt.edu; wang.xq@sz.tsinghua.edu.cn; \{mathshenli,dacheng.tao\}@gmail.com
}
}

\begin{document}

\maketitle

\begin{abstract}
  Personalized federated learning (PFL) aims to produce the greatest personalized model for each client to face an insurmountable problem -- data heterogeneity in real FL systems. However, almost all existing works have to face large communication burdens and the risk of disruption if the central server fails. Only limited efforts have been used in a decentralized way but still suffers from inferior representation ability due to sharing the full model with its neighbors. Therefore, in this paper, we propose a personalized FL framework with a decentralized partial model training called DFedAlt. It personalizes the “right” components in the modern deep models by alternately updating the shared and personal parameters to train partially personalized models in a peer-to-peer manner. To further 
  promote the shared parameters aggregation process, we propose DFedSalt integrating the local Sharpness Aware Minimization (SAM) optimizer to update the shared parameters. It adds proper perturbation in the direction of the gradient to overcome the shared model inconsistency across clients. Theoretically, we provide convergence analysis of both algorithms in the general non-convex setting for decentralized partial model training in PFL. Our experiments on several real-world data with various data partition settings demonstrate that (i) decentralized training is more suitable for partial personalization, which results in state-of-the-art (SOTA) accuracy compared with the SOTA PFL baselines; (ii) the shared parameters with proper perturbation make partial personalized FL more suitable for decentralized training, where DFedSalt achieves most competitive performance.

\end{abstract}

\section{Introduction}

\begin{wrapfigure}[11]{r}{0.58\linewidth}
    \centering
     \vspace{-0.5cm}
    \begin{subfigure}{1.0\linewidth}
		\centering
          \includegraphics[width=1.0\textwidth]{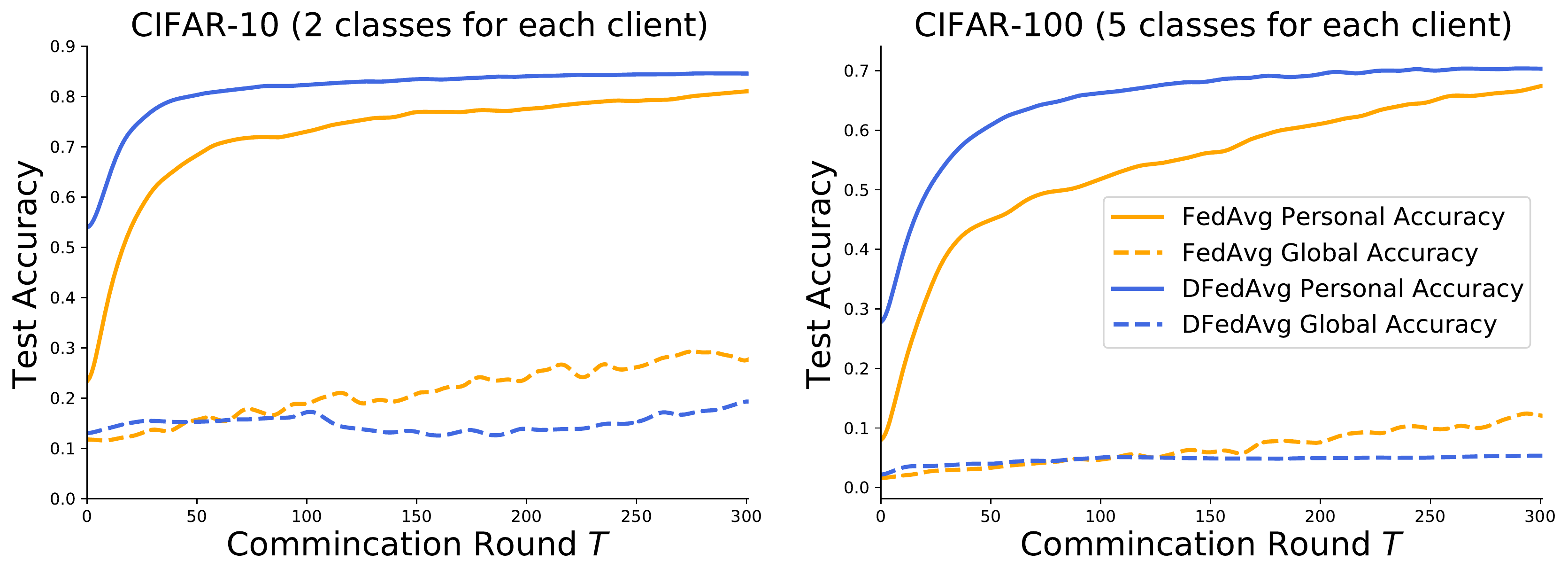}
	\end{subfigure}
\caption{\small The training progress comparison between FedAvg and DFedAvg. DFedAvg can achieve competitive personal accuracy with fewer rounds than FedAvg compared with global accuracy.}
\label{fig:motivation}
\end{wrapfigure}
To solve a major challenge --- the data heterogeneity problem, most works propose to achieve many personalized individual models for all clients rather than a single global model fitting the whole data distribution from all clients, called personalized federated learning (PFL) \cite{Kairouz2021Advances}. However, almost all existing works suffer from communication burdens and the risk of disruption if the central server fails in the centralized FL (CFL) setting \cite{beltran2022decentralized,Kairouz2021Advances}.
Only limited efforts focusing on these issues have been used in a decentralized way for model aggregation, called the decentralized FL (DFL) setting. For instance, the works \cite{jeong2023personalized,sadiev2022decentralized,Rong2022DisPFL} 
leverage the full personalized model to communicate with the neighbors for each client in a peer-to-peer manner. 
%\begin{figure*}[ht]
%\centering
%\includegraphics[width=0.6\textwidth]{images/motivation_cifar10.pdf}
%\centering
% \vspace{-0.35cm}
%\caption{\small The global and the personal accuracy between Fedavg and DFedavg in CIFAR-10 and CIFAR-100 with various heterogenous data partitions. \ls{we may merely put two figures in the introduction. The results of CIFAR 100 can be placed in the appendix}}
% \vspace{-0.35cm}
%\label{fig:abla}
%\end{figure*}

\textbf{Motivation.} Firstly, we comprehensively investigate the role of decentralized training in personalized FL (PFL) by conducting some toy experiments for FedAvg \cite{mcmahan2017communication} and decentralized FedAvg (DFedAvg) \cite{sun2022decentralized} on both CIFAR-10 and CIFAR-100 datasets with a pathological partition approach in Figure \ref{fig:motivation}, in which data distributions are very heterogeneous across clients --- only $2$ classes on CIFAR-10 and $5$ classes on CIFAR-100 for each client, and a very sparse topology --- Ring topology is applied on DFedAvg with $100$ clients. It is clear that for one global model accuracy, the performance of DFL is worse than that of CFL. In contrast, personal accuracy in DFL is better than that in CFL.  It indicates that the decentralized training method might be more suitable for PFL tasks. 
However, existing works focusing on a decentralized way in PFL \cite{Rong2022DisPFL,sadiev2022decentralized,jeong2023personalized}, still face an inferior representation ability challenge in modern deep models due to losing unique information in each client, which is caused by full model aggregation with its neighbors \cite{pillutla2022federated}. 
Existing work in PFL \cite{pillutla2022federated} also provides a similar opinion --- a fully personalized model may lead to “catastrophic forgetting” \cite{mccloskey1989catastrophic}. Therefore, a question arises: 
\begin{tcolorbox}[notitle, sharp corners, colframe=black, colback=white, 
       boxrule=1pt, boxsep=0pt, enhanced, shadow={3pt}{-3pt}{0pt}{opacity=1,black}]
\begin{center}
  \emph{\small Can we seek out suitable personalization in FL via decentralized partial model training?} 
\end{center}
\end{tcolorbox}

% \iffalse
% \begin{wrapfigure}[22]{r}{0.35\linewidth}
%     \centering
%      \vspace{-0.4cm}
%     \begin{subfigure}{0.8\linewidth}
% 		\centering
% 		\includegraphics[width=1.0\linewidth]{images/dsgd-top-dir.pdf}
% 		\caption{Dirichlet-0.3}
% 	    \label{Dirichlet 0.3}
% 	\end{subfigure}
%         % \vspace{-0.4cm}
%          \begin{subfigure}{0.8\linewidth}
% 		\centering
% 		\includegraphics[width=1.0\linewidth]{images/dsgd-top-pat.pdf}
% 		\caption{Pathlogical-2}
% 	    \label{Pathlogical 2}
% 	\end{subfigure}
%         % \vspace{-0.2cm}
%         \caption{\small Personal test accuracy of DFedAvg in various communication topologies with Dirichlet partition (a) and pathological partition (b) on CIFAR-10. \ls{this figure can not reflect the motivation of DFedSalt, we can move it to somewhere else.}}
% 	\label{da_chutian}
%     % \vspace{-10.5cm}
% \end{wrapfigure}
% \fi

\medskip
\textbf{Contributions.} To seek out a better solution for PFL via sharing the partial “right” components selected with domain knowledge and unsharing the “unique” information in each client, we adopt the decentralized partial model training in PFL, named DFedAlt, which decomposes each local model as shared part and personalized part and then optimizes them alternatively in a decentralized manner. To the best of our knowledge, we are the first to explore decentralized partial personalization in PFL and face this inferior representation ability challenge by decomposing each local model and only averaging a shared part with its neighbors of each client.
%As shown in Figure \ref{da_chutian}, we can see that various communication topologies have a great influence on the performance and convergence speed of the personalized model. Therefore, To make the DFL setting more suitable for partial personalization, we integrate the local SAM optimizer into the shared parameters update to search for the shared parameters with uniformly low loss values, dubbed as DFedSalt. \ls{this part should be reorganized. The logic in enhancing DFedAlt as DFedSalt should be reconsidered.} 
Furthermore, we propose an enhanced version of DFedAlt, called DFedSalt, which integrates local SAM optimizer \cite{foret2021sharpnessaware} to update the shared parameters. Specifically, it searches for the shared parameters with uniformly low loss values by adding proper perturbation in the direction of the gradient, thereby promoting the process of local model aggregations in each client (see \textbf{Section} \ref{algo}). Meanwhile, we present the non-trivially theoretical analysis for both DFedAlt and DFedSalt algorithms in the general non-convex setting (see \textbf{Section} \ref{theory}), which can analyze the ill impact of the statistical heterogeneity and smoothness of loss functions on
the convergence with partial personalization and SAM optimizer for the shared model. Empirically, we conduct extensive experiments on CIFAR-10, CIFAR-100, and Tiny-ImageNet datasets in non-IID settings with different data partitions, such as Dirichlet settings with various $\alpha$ and pathological settings with various limited classes in each client. Experimental results confirm that our algorithms can achieve competitive performance relative to many SOTA PFL baselines (see \textbf{Section} \ref{exper}).

% \vspace{0.3cm}

% \textbf{Technical Difficulty.}

In summary, we provide a comprehensive study focusing specifically on decentralized partial model personalization in PFL. Our main contributions lie in four-fold:
\begin{itemize}
    \item We seek out suitable personalization in PFL with decentralized training and propose DFedAlt  via alternately updating the shared part and personal part in a peer-to-peer manner.
    \item To further overcome the model inconsistency of the shared parameters, we propose DFedSalt, which enhances DFedAlt by integrating the local SAM optimizer into the shared parameters.
    \item We provide \emph{convergence guarantees} for the DFedAlt and DFedSalt methods in the general non-convex setting with \emph{decentralized partial participation} in PFL.
    \item We conduct \emph{extensive experiments} on realistic data tasks with various data partition ways, evaluating the efficacy of our algorithms compared with some SOTA PFL baselines.
\end{itemize}
\section{Related Work}
\textbf{Personalized Federated Learning (PFL).}
% Compared to the FL pursuing a more robust global model for clients’ non-iid distributions, the PFL aims to produce the greatest personalized models for each client. From the perspective of learning personalized models, 
In recent years, the research works in PFL can be roughly divided into four categories:  parameter decoupling \cite{arivazhagan2019federated, collins2021exploiting, oh2021fedbabu}, knowledge distillation \cite{li2019fedmd, lin2020ensemble, he2020group}, multi-task learning \cite{huang2021personalized, shoham2019overcoming}, model interpolation \cite{deng2020adaptive, diao2020heterofl} and clustering \cite{ghosh2020efficient, sattler2020clustered}. For instance, Ditto \cite{li2021ditto} adds a regularization term to address simultaneously robustness and fairness constraints in PFL for federated multi-task learning. Recently, Fed-RoD \cite{chen2021bridging} leverages a global body and two heads, e.g., the generic head trained with class-balanced loss and the personalized head trained with empirical loss, to generate both great generic performance and personalized performance. 
More details can be referred to in \cite{tan2022towards}. In this paper, we mainly focus on the parameter decoupling methods, which divide the model into a global shared part and a personalized part, also called \emph{partial personalization}. 
% For a fair comparison and exploration, we mainly present the related works about partial personalization and full personalization. The latter represents the whole model as the personalized model without decoupling parameters. 

\textbf{Partial Personalization in FL.} 
Existing works demonstrate that partial personalization often outperforms full personalization. Specifically, FedPer \cite{arivazhagan2019federated} uses the one global body with many local heads approach and only shares the body layers with the server. After that, the entire model jointly is learned for each local client.
FedRep \cite{collins2021exploiting} learns the entire model sequentially with the head updating first and the body later, and only shares the body layers with the server too.   Where the linear convergence is also presented for a two-layer linear network. FedBABU \cite{oh2021fedbabu} trains the global body with a fixed head for all clients and finally fine-tunes the personalized heads on the basis of the consensus body. And they \citep{oh2021fedbabu} also explore empirically that mixing heads in heterogeneous scenarios will lead to the performance degradation of local models. 
FedSim and FedAlt are proposed in \cite{pillutla2022federated}, they provide the first convergence analyses of both algorithms in the general nonconvex setting with partial participation, where FedAlt leverages the alternating update algorithm similar to FedRep, and FedSim uses the simultaneous update algorithm similar to FedPer.
% The global body and the generic head are aggregates in the server for better generic performance while the personalized head is kept locally for better personalized performance.

% About the \textbf{Full personalization} methods, 

% \begin{itemize}
%     \item \textbf{Partial personalization.}
%     \item \textbf{Full personalization.}
% \end{itemize}

\textbf{Decentralized Federated Learning (DFL).}
Due to the participants having different hardware and network capabilities in the real federated system, DFL is an encouraging field of research that has repeatedly been reported as challenging in several review articles in recent years \cite{beltran2022decentralized,kang2022blockchain,LiJunBlockchain,nguyen2022novel,wang2022accelerating,wang2020learning,yu2020proactive}.
In DFL, the clients only connect with their neighbors and its goal is to make all local models tend to a unified model through peer-to-peer communication. For some applications, 
BrainTorrent \cite{roy2019braintorrent} is the first serverless, peer-to-peer FL approach and applied to medical applications in a highly dynamic peer-to-peer FL environment.
Similar to general FL methods such as \cite{mcmahan2017communication}, we discuss the DFL methods considering both multi-steps local iterations and various communication topologies.\footnote{In decentralized/distributed training, they also focus on peer-to-peer communication, but one-step local iteration is adopted, due to the gradient computation being more focused than the communication burden. More detailed related works in decentralized/distributed training are placed in \textbf{Appendix} \ref{ap:related_works}.} Specifically, DFedAvg \cite{sun2022decentralized} applies the multiple local iterations with SGD and quantization method to reduce the communication cost and provide the convergence results in various convex settings. DisPFL \cite{Rong2022DisPFL} customizes the personalized model and pruned mask for each client to further lower the communication and computation cost. KD-PDFL \cite{jeong2023personalized} leverages knowledge distillation technique to empower each device so as to discern statistical distances between local models. The work in  \cite{sadiev2022decentralized} presents lower bounds on the communication and local computation costs for this personalized FL formulation in a peer-to-peer manner.
DFedSAM \cite{shi2023improving} integrates Sharpness Awareness Minimization (SAM) into DFL to improve the model consistency across clients.

The most related works to this paper lie in (i) CFL methods: FedPer \cite{arivazhagan2019federated}, FedRep \cite{collins2021exploiting}, FedBABU \cite{oh2021fedbabu}, Ditto \cite{li2021ditto}, and Fed-RoD\cite{chen2021bridging}; (ii) DFL methods: DFedAvgM \cite{sun2022decentralized}, Dis-PFL \cite{Rong2022DisPFL}, and DFedSAM \cite{shi2023improving}.
However, in PFL, almost all existing works have to face large communication burdens and the risk of disruption if the central server fails.
Only limited efforts have been used in a decentralized way but still suffers from  poor representation ability due to sharing the full personalized model with its neighbors. 
Therefore,
different from existing works, we try to seek out suitable personalization in FL via decentralized partial model training. Meanwhile, we provide the first convergence analysis on decentralized partial model personalization in FL, which is non-trivial. %Consequently, we conduct extensive experiments on real data to evaluate the efficacy of our algorithms.

\section{Methodology}\label{algo}

In this section, we define the problem setup for DFL and decentralized partial personalized models in PFL at first. After that, we present two algorithms: DFedAlt and DFedSalt in PFL, which leverages the decentralized partial model personalization technique to generate better representation ability while achieving SOTA performance relative to many related PFL methods.

\subsection{Problem Setup} 

\textbf{Decentralized Federated Learning (DFL).}
We consider a typical setting of DFL with~$m$ clients, where each client~$i$ has the data distribution $\mathcal{D}_i$. 
Let $w\in\mathbb{R}^d$ represent the parameters of a machine learning model and $F_i(w;\xi)$ is the local objective function associated with the training data samples $\xi$.
Then the loss function associated with client~$i$ is 
$F_i(w)=\mathbb{E}_{\xi\sim \mathcal{D}_i} F_i(w;\xi)$.
After that, a common objective of DFL is the following finite-sum stochastic non-convex minimization problem:\\
\begin{equation}\label{dec}
    \small \min_{w\in \mathbb{R}^d} F(w):=\frac{1}{m}\sum_{i=1}^m F_i(w).
\end{equation}
In the decentralized network topology, the communication between clients can be modeled as an undirected connected graph $\mathcal{G} = (\mathcal{N},\mathcal{V}, \boldsymbol{W}  )$, where $\mathcal{N} = \{1, 2, \ldots, m\}$ represents the set of clients, $\mathcal{V} \subseteq  \mathcal{N} \times  \mathcal{N}$ represents the set of communication channels, each connecting two distinct clients, and the gossip/mixing matrix $\boldsymbol{W}$ records whether the communication connects or not between any two clients. As below, we present the definition of $\boldsymbol{W}$:
\begin{definition}[The gossip/mixing matrix \cite{sun2022decentralized}] 
\label{def:gossip_matrix}
The gossip matrix $\small {\bf W} = [w_{i,j}] \in [0,1]^{m\times m}$  is assumed to have these properties:
(i) (Graph) If $\small i\neq j$ and $(i,j) \notin {\cal V}$, then $w_{i,j} =0$, otherwise, $w_{i,j} >0$;
(ii) (Symmetry) $\small {\bf W} = {\bf W}^{\top}$;
(iii) (Null space property) $\small \mathrm{null} \{{\bf I}-{\bf W}\} = \mathrm{span}\{\bf 1\}$;
(iv) (Spectral property) $\small {\bf I} \succeq {\bf W} \succ -{\bf I}$.  \emph{Under these properties, the eigenvalues of ${\bf W}$ satisfies $\small 1=|\lambda_1({\bf W)})|> |\lambda_2({\bf W)})| \ge \dots \ge |\lambda_m({\bf W)})|$. And $\lambda:=\max\{|\lambda_2({\bf W)}|,|\lambda_m({\bf W)})|\}$ and $1-\lambda \in (0,1]$ is the spectral gap of $\bf W$, which usually measures the degree of the network topology.}
\end{definition}

% Furthermore, 
% In addition, we assume that Equation (\ref{dec}) is well-defined and denote $f^{*}$ as the minimal value of $f$: $f(x)\ge f(x^*)=f^*$ for all $x\in \mathbb{R}^{d}$. 

% \noindent

% \emph{1. Various communication topologies.} The topology is measured by the spectral gap $1-\lambda \in (0,1]$ of $\bf W$, and the value of $\lambda$ increases as the connectivity is more sparse. It has a great ill-impact on model training (convergence rate and generalization ability), especially on heterogeneous data or in face of the sparse connectivity of communication networks \cite{sun2022decentralized, zhu2022topology, shi2023improving}, e.g., Ring topology and Grid topology where $\lambda \approx 16 \pi^2/(3m^2)$  and $\lambda =\mathcal{O}(1/(m\log_2(m)))$, $m$ is the total client's size in Table 1 of \cite{zhu2022topology}, respectively.
% % \noindent

% \emph{2. Multi-step local iterations.} After multiple local iterations, the gradient estimation may be failed to be unbiased. That means the theoretical analysis and empirical efficacy may be more difficult than the one-step local iteration. \footnote{One-step local iteration is often used in decentralized/distributed training \cite{lian2017can,zhu2022topology,ye2020decentralized,Hashemi2022On}.}

\textbf{DFL with Partial Personalized Models.}  
Below, we present a general setting of DFL with \emph{partial model personalization} for considering the communication overhead.
Specifically, the model parameters are partitioned into two parts: 
the \emph{shared} parameters $u\in\mathbb{R}^{d_0}$ and the \emph{personal}
parameters $v_i\in\mathbb{R}^{d_i}$ for $i=1,\ldots,m$. 
The full model on client~$i$ is denoted as $w_i=(u_i,v_i)$. To simplify presentation, we denote $V=(v_1,\ldots,v_m)\in\mathbb{R}^{d_1+\ldots+d_m}$, and then our goal is to solve this problem:
\begin{equation}\label{eqn:pfl-partial}
\min_{u, V} \quad F(u, V):=\frac{1}{m} \sum_{i=1}^m F_i\left(u, v_i\right),
\end{equation}
where $u$ denotes the consensus model averaged with all shared models $u_i$, that is $u = \frac{1}{m}\sum_{i=1}^m u_i$.
Moreover, we consider the more general non-convex setting with local functions $F_i\left(u_i, v_i\right)=  {E}_{\xi_i \sim \mathcal{D}_i}\left[F_i\left(u_i, v_i; \xi_i\right)\right]$, and for brevity, also use ${\nabla}_u$ and ${\nabla}_v$ to represent stochastic gradients with respect to~$u_i$ and $v_i$, respectively.

In the DFL setting, the shared parameters $u_i$ of each client $i$ are sent out to the neighbors of client $i$ from the neighborhood set with adjacency matrix $\mathbf{W}$, which records the communication connections between any two clients (communication topology). In contrast, the personal parameters $v_i$ only perform multiple local iterations in each client $i$ and do not be sent out.

%
% In {LocalRep}, the personal parameters are updated first with the received shared parameters fixed, then the shared parameters are updated with the new personal parameters fixed.
\begin{figure*}[t]
\centering
\includegraphics[width=0.99\textwidth]{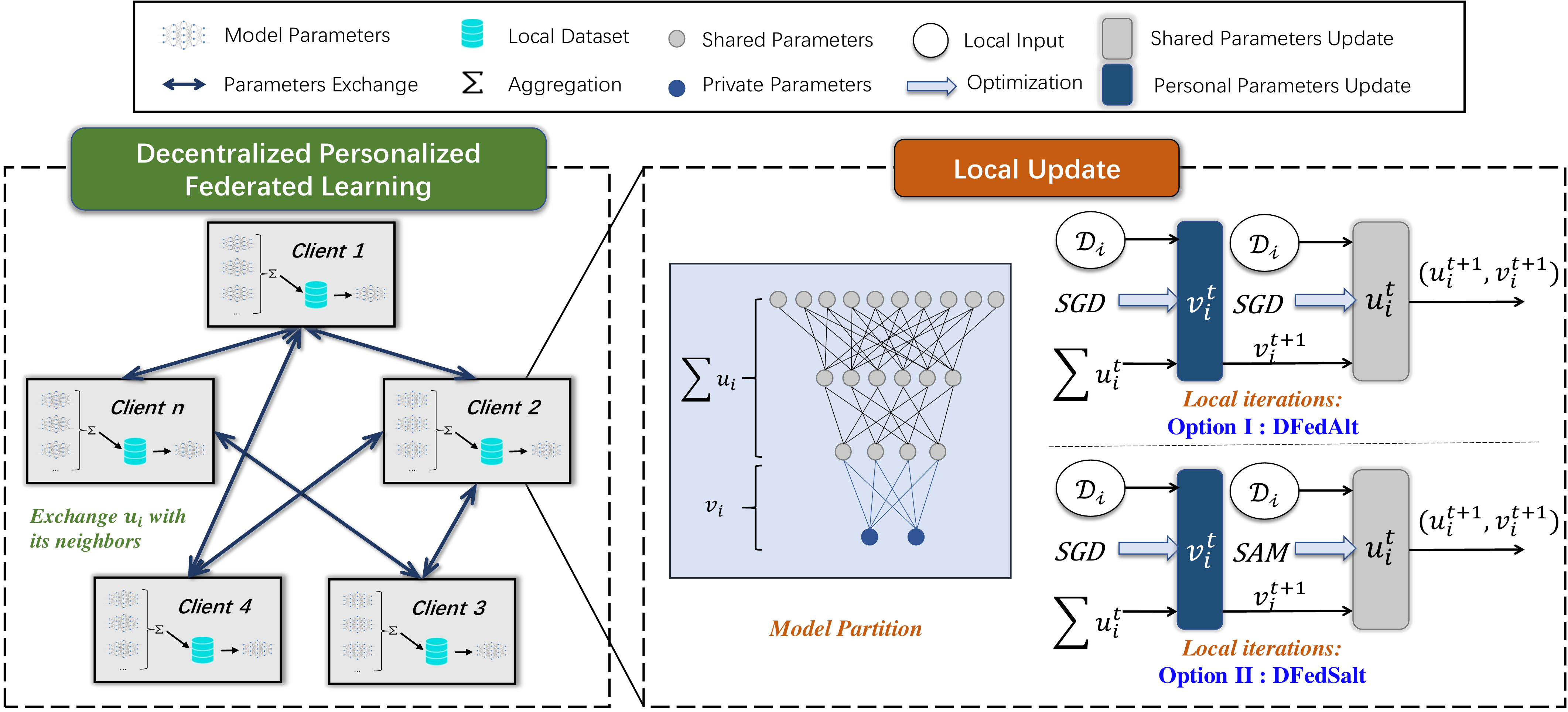}
\centering
% \vspace{-0.35cm}
\caption{\small An overview of the proposed DFedAlt and DFedSalt frameworks. }
 \vspace{-0.35cm}
\label{fig:abla}
\end{figure*}

\subsection{DFedAlt and DFedSalt Algorithms}

In this subsection, we present the DFedAlt and DFedSalt algorithms for solving problem~\eqref{eqn:pfl-partial}. The detailed procedure and pipeline are presented in Algorithm \ref{DFedSAlt} and Figure \ref{fig:abla}, respectively.

\textbf{DFedAlt.} To explore the possible partial personalization benefit of DFL, we present a useful partial personalization framework in DFL, named DFedAlt, which leverages the alternating update approach for model training. Specifically, the personal parameters $v_i$ for each client perform multiple local iterations at first in line 6. After that, the shared parameters $u_i$ perform multiple local iterations in line 10. After multiple local iterations of shared parameters $u_i$ in each client $i$, the resulting parameters  $  {z}^t_{i} \gets    {u}^{t,K_u}_{i}$ is sent to its neighbors in line 12. And then each client updates its shared parameters by averaging its neighbors' shared parameters (including itself).

\begin{wrapfigure}[30]{r}{0.58\textwidth}
\hspace{0.1cm}
\begin{minipage}{0.58\textwidth}
\vspace{-0.1cm}
\begin{algorithm}[H]
\small
\caption{DFedAlt \color{black}{and} DFedSalt}
\label{DFedSAlt}
% \centering
\SetKwData{Left}{left}\SetKwData{This}{this}\SetKwData{Up}{up} \SetKwFunction{Union}{Union}\SetKwFunction{FindCompress}{FindCompress}
\SetKwInOut{Input}{Input}\SetKwInOut{Output}{Output}

\Input{Total number of devices $m$, total number of communication rounds $T$,  local learning rate $\eta_{u}$ and $\eta_{v}$, total number of local iterates $K_{u}$ and $K_{v}$.} 

\Output{Personalized model $  {u}^T_{i}$ and $  {v}^T_{i}$ after the final communication of all clients.}

\textbf{Initialization:}  Randomly initialize each device's shared parameters $  {u}^{0}_{i}$ and personal parameters $  {v}^{0}_{i}$.

\For{$t=0$ \KwTo $T-1$}{
    \For{client $i$ in parallel }{
        Set $  {u}^{t,0}_{i} \gets    {u}^t_{i}$ and sample a batch of local data $\xi_i$ and calculate local gradient iteration.\\
        \For{$k=0$ \KwTo $K_{v}-1$ }{  
        Perform personal parameters $  {v}_i$ update: 
        ${  {v}_{i}^{t, k + 1}} = {  {v}_{i}^{t, k}} - {\eta _v}{\nabla _v}F_i({u}^{ t,0}_{i},{  {v}_{i}^{t,k}};{\xi _i})$.}

        $  {v}_{i}^{t + 1} \gets   {v}_{i}^{t, K_v}$. 
        
        \For{$k=0$ \KwTo $K_{u}-1$ }{
        % \colorbox[rgb]{1.0, 0.55, 0.41}{
        Update shared parameters $  {u}_i$ via \color{blue}{Option I} \color{black}{or} \color{blue}{II}.
        % \texttt{// DFedSalt}\\
        }
        $  {z}^t_{i} \gets    {u}^{t,K_u}_{i}.$ 
        Receive neighbors’ shared models $ {z}_{j}^{t}$ with adjacency matrix $\boldsymbol{W}$:
 $  {u}^{t+1}_{i} = \sum_{l \in \mathcal{N}(i) } w_{i,l}  {z}^{t}_{i}$.}}
% \vspace{-0.1in}
% \rule[-10pt]{14cm}{0.03em}
%\begin{multicols}{2}
\color{blue}{Option I: (DFedAlt)} Find a minimum for ${u}_i$  with SGD\\
\color{black}{ ${  {u}_{i}^{t, k + 1}} =   {u}_{i}^{t, k} - {\eta _u}\nabla_u F_i({u}_{i}^{t, k}, {v}_{i}^{t + 1};\xi_i)$.}

{\color{blue}{Option II: (DFedSalt)}. Find a flat minimum for ${u}_i$  with SAM}

$\epsilon({u}_{i}^{t, k})=\rho \frac{\nabla_u F_i({u}_{i}^{t, k}, {v}_{i}^{t + 1};\xi_i)}{\| \nabla_u F_i({u}_{i}^{t, k}, {v}_{i}^{t + 1};\xi_i)  \|_2} $.

${  {u}_{i}^{t, k + 1}} =   {u}_{i}^{t, k} - {\eta _u}\nabla_u F_i({u}_{i}^{t, k} + \epsilon({u}_{i}^{t, k}), {v}_{i}^{t + 1};\xi_i)$.
%\end{multicols}
\end{algorithm}
\end{minipage}
\end{wrapfigure}
\textbf{An Enhanced Algorithm: DFedSalt.}
In FL, the model inconsistency issue is a major challenge across clients due to data heterogeneity \cite{shi2023improving,sun2022decentralized}, resulting in severe over-fitting of local models. In particular, sparse communication topology is also a key factor causing this issue in DFL \cite{shi2023improving}. 
Therefore, to further make partial personalization more suitable for DFL by decreasing the generalization error of shared parameters, we propose DFedSalt, which integrates the SAM optimizer into the local iteration update of shared parameters $u_i$.  Specifically, we adopt proper perturbation in the direction of the local gradient of the shared parameters $u_i$. At first, the gradient $\nabla_u F_i({u}_{i}^{t, k}, {v}_{i}^{t + 1};\xi_i)$ of $u_i$ is calculated on mini-batch data $\xi_i$ for each client $i$. And then, we calculate the perturbation value in line 18, where $\rho$ is a hyper-parameter for controlling the value of the perturbation radius. Finally, adding the perturbation term into the direction of gradient $\nabla_u F_i({u}_{i}^{t, k}, {v}_{i}^{t + 1};\xi_i)$ in line 18. The local averaging of $u_i$ is the same as the DFedAlt algorithm.

\section{Theoretical Analysis}\label{theory}
In this section, we present the convergence analysis in DFedAlt and DFedSalt methods for the characterization of convergence speed and the exploration of how partial personalization and SAM optimizer work. Below, we state some general assumptions at first.
% \begin{definition}\label{noniid_para}
% \textbf{(Homogeneity parameter).}  [Definition 2, \cite{li2020communication}] For any $i \in \{1,2,\ldots,m\}$ and the parameter $\mathbf{x} \in \mathbb{R}^d$, the homogeneity parameter $\beta$ can be defined as:
% \begin{equation}
% \small
%     \beta := \max_{1 \leq i \leq m} \beta_i,~~~ with ~\beta_i := \sup_{\mathbf{x} \in \mathbb{R}^d}\left \| \nabla f_i(\mathbf{x}) - \nabla f(\mathbf{x}) \right \|. \nonumber
% \end{equation}
% \end{definition}

\begin{assumption}[Smoothness \cite{pillutla2022federated}]\label{assmp:smoothness}
For each client $i=\{1,\ldots,m\}$, the function $F_i$ is continuously differentiable.  
There exist constants $L_u, L_v, L_{uv}, L_{vu}$ such that
for each client $i=\{1,\ldots,m\}$:
\begin{itemize}[topsep=0pt,itemsep=0pt,leftmargin=\widthof{(a)}]
\item $\nabla_u F_i(u_i,v_i)$ is $L_u$--Lipschitz with respect to~$u_i$ and $L_{uv}$--Lipschitz with respect to~$v_i$
\item $\nabla_v F_i(u_i,v_i)$ is $L_v$--Lipschitz with respect to~$v_i$ and $L_{vu}$--Lipschitz with respect to~$u_i$.
\end{itemize}
We summarize the relative cross-sensitivity of $\nabla_u F_i$ with respect to~$v_i$ and $\nabla_v F_i$ with respect to~$u$ with 
the scalar
\begin{equation}\label{eqn:chi-def}
\chi := \max\{L_{uv},\,L_{vu}\}\big/\sqrt{L_u L_v}.\nonumber
\end{equation}
\end{assumption}

\begin{assumption}[Bounded Variance \cite{pillutla2022federated}] \label{assmp:stoc-grad-var}
The stochastic gradients in Algorithm \ref{DFedSAlt} have bounded variance. That is, for all $u_i$ and $v_i$, 
% \begin{align*}
%     \E\bigl[ \widetilde\nabla_u F_i(u_i, v_i)\bigr] &= \nabla_u F_i(u_i, v_i),
%     %
%     \\
%     \E\bigl[ \widetilde\nabla_v F_i (u_i, v_i)\bigr] &= \nabla_v F_i(u_i, v_i)\,.
% \end{align*}
there exist constants $\sigma_u$ and $\sigma_v$ such that
\begin{align*}
    \E\bigl[\bigl\|\nabla_u F_i(u_i, v_i; \xi_i) - \nabla_u F_i(u_i, v_i)\bigr\|^2\bigr] &\le \sigma_u^2,~ 
%    \\ %
    \E\bigl[\bigl\|\nabla_v F_i(u_i, v_i; \xi_i) - \nabla_v F_i(u_i, v_i)\bigr\|^2\bigr] &\le \sigma_v^2 \,.
\end{align*}
\end{assumption}

\begin{assumption}[Partial Gradient Diversity \cite{pillutla2022federated}]
\label{assmp:grad-diversity.}
There exist a constant $\delta\ge 0$ such that
\[
\textstyle
    \frac{1}{m}\sum_{i=1}^m \bigl\|\nabla_u F_i(u_i, v_i) - \nabla_u F(u_i, V)\bigr\|^2
    \le \delta^2,~ \forall u_i,~V.
\]
\end{assumption}

The above assumptions are mild and commonly used in the convergence analysis of FL \cite{sun2022decentralized,shi2023improving,ghadimi2013stochastic,yang2021achieving,bottou2018optimization,reddi2020adaptive, huang2022achieving, Qu2022Generalized}. 
% Note that we also adopt the average point of all clients' parameters $\bar{u}^t = \frac{1}{m}\sum_{i=1}^m u^t_i$ to characterize the convergence of DFedAlt and DFedSalt.

\textbf{About the Challenges of Convergence Analysis.}  Due to the central server being discarded, various communication connections will become an important factor for decentralized optimization. Furthermore, communication is more careful in general classical FL scenarios rather than computation \cite{mcmahan2017communication, Li2020fl, Kairouz2021Advances, Qu2022Generalized}. So the client adopts multi-step local iterations such as FedAvg \cite{mcmahan2017communication}, which may lead to the local gradient failing to be unbiased. Because of these factors, technical difficulty exists in our theoretical analysis. How to analyze the convergence of decomposed model parameters while delivering the impact of communication topology.
In this paper, we adopt the averaged shared parameter $\bar{u}^t \!=\! \frac{1}{m}\sum^m_{i=1} u^t_i$ of all clients to be the approximated solution of problem ~\eqref{eqn:pfl-partial} due to only the shared parameters being communicated with the neighbors \cite{sun2022decentralized,shi2023improving}.
Now, we present the rigorous convergence rate of DFedAlt and DFedSalt algorithms as follows.
\begin{theorem}[Convergence Analysis for DFedAlt]\label{th:conver_alt1}
Under assumptions 1-3 and definition 1, the local learning rates satisfy $\eta_u=\mathcal{O}({1}/{L_uK_u\sqrt{T}}), \eta_v=\mathcal{O}({1}/{L_vK_v\sqrt{T}})$, $F^{*}$ is denoted as the minimal value of $F$, i.e., $F(\bar{u}, V)\ge F^*$ for all $\bar{u} \in \mathbb{R}^{d}$, and $V=(v_1,\ldots,v_m)\in\mathbb{R}^{d_1+\ldots+d_m}$. Let $\bar{u}^t = \frac{1}{m}\sum_{i=1}^m u^t_i$ and denote $\Delta_{\bar{u}}^t$ and $\Delta_{v}^t$ as:
\begin{center}
    $\Delta_{\bar{u}}^t = \bigl \| \nabla_u F(\bar{u}^t, V^{t+1}) \bigr\|^2 \,,  \quad \text{and} ~~~
        \Delta_v^t = \frac{1}{m}\sum_{i=1}^m \bigl\|\nabla_v F_i(u_i^t, v_i^t)\bigr\|^2 \,.$
\end{center}

% \begin{align}
%    \Delta_{\bar{u}}^t = \bigl \| \nabla_u F(\bar{u}^t, V^{t+1}) \bigr\|^2 \,,  \quad \text{and} ~~~
%         \Delta_v^t = \frac{1}{m}\sum_{i=1}^m \bigl\|\nabla_v F_i(u_i^t, v_i^t)\bigr\|^2 \,. \nonumber
% \end{align}
Therefore, we have the convergence rate as below:
\begin{equation} \label{th:conver_alt}  
\small
\frac{1}{T}\sum_{i=1}^T \bigl(\frac{1}{L_u} \E \bigl[\Delta_{\bar{u}}^t \bigr] + \frac{1}{L_v} \E [\Delta_{v}^t \bigr] \bigr) \leq \mathcal{O}\Big(\frac{F(\bar{u}^1, V^1) - F^*}{\sqrt{T}} + \frac{\sigma_1^2}{\sqrt{T}} + \frac{\sigma_2^2}{T(1-\lambda)^2} \Big),
\end{equation}
where     
% \begin{center}
%     $\sigma_1^2 = \frac{\sigma_u^2 + \delta^2}{L_u}\, ,~~~
%         \sigma_2^2 = \frac{\sigma_v^2}{L_v} +\frac{\chi^2L_v(\sigma_u^2+\delta^2)}{L_u}\, .$
% \end{center}
\begin{align}
\small
        \sigma_1^2 = \frac{\sigma_v^2(L_v+1)}{L_v^2} + \frac{L_{vu}^2(\sigma_u^2+\delta^2)}{L_u^2} = \frac{\sigma_v^2(L_v+1)}{L_v^2} + \frac{\chi^2L_v(\sigma_u^2+\delta^2)}{L_u} \, ,~~~
        \sigma_2^2 = \frac{\sigma_u^2+\delta^2}{L_u}\, .\nonumber
\end{align}

\begin{remark}
\emph{These variables have a significant influence on the convergence bound. Specifically, measuring the statistical heterogeneity, such as local variance $\sigma_u^2, \sigma_v^2$ and global diversity, the smoothness of local loss functions such as $L_u$, $L_v$, and $L_{vu}$, and the communication topology measured by $1-\lambda$.}
\end{remark}
 
\end{theorem}

\begin{theorem}[Convergence Analysis for DFedSalt]\label{th:conver_salt1}
Under assumptions 1-3 and definition 1, the local learning rates satisfy $\eta_u=\mathcal{O}({1}/{L_uK_u\sqrt{T}}), \eta_v=\mathcal{O}({1}/{L_vK_v\sqrt{T}})$. Let $\bar{u}^t = \frac{1}{m}\sum_{i=1}^m u^t_i$ and denote $\Delta_{\bar{u}}^t$ and $\Delta_{v}^t$ as Theorem \ref{th:conver_alt1}.
% \begin{align}
%     \Delta_{\bar{u}}^t = \bigl \| \nabla_u F(\bar{u}^t, V^{t+1}) \bigr\|^2 \,,  \quad \text{and} ~~~
%         \Delta_v^t = \frac{1}{m}\sum_{i=1}^m \bigl\|\nabla_v F_i(u_i^t, v_i^t)\bigr\|^2 \,. \nonumber
% \end{align}
When the perturbation amplitude $\rho$ is
proportional to the learning rate, e.g., $\rho = \mathcal{O}(1/\sqrt{T})$,
the sequence of outputs $\Delta_{\bar{u}}^t $ and $\Delta_{v}^t$ generated by DFedSalt, we have:
% Therefore, we have the convergence rate as below:
\begin{equation} \label{th:conver_salt}  
\small
\frac{1}{T}\sum_{i=1}^T \bigl(\frac{1}{L_u} \E \bigl[\Delta_{\bar{u}}^t \bigr] \!+\! \frac{1}{L_v} \E [\Delta_{v}^t \bigr] \bigr) \!\leq\! \mathcal{O}\Big(\frac{F(\bar{u}^1, V^1) \!-\! F^*}{\sqrt{T}} \!+ \frac{\sigma_v^2(L_v+1)}{L_v^2\sqrt{T}} + \! \frac{L_u}{T} + \frac{\sigma^2L^2_{vu}}{T^{1/2}(1\!-\!\lambda)^2}  + \frac{\sigma^2L_u}{T(1\!-\!\lambda)^2} \Big),
\end{equation}
where $\small \mathcal{O}\Big(\sigma^2 \Big) = \mathcal{O}\Big( \frac{\rho^2}{K_u} + \frac{\sigma_u^2+\delta^2}{L_u^2}\Big) =  \mathcal{O}\Big(\frac{1}{K_uT} + \frac{\sigma_u^2+\delta^2}{L_u^2} \Big)\,$ when $\rho = \mathcal{O}(\frac{1}{\sqrt{T}})$.
\begin{remark}
\emph{It is clear that the bound is facilitated via SAM optimizer from the smoothness-enabled perspective, such as $L_u^2$ and $L_{vu}^2$. Thus, the shared model $u_i$ may be flatter, thereby decreasing the generalization error of the whole model $w_i=(u_i, v_i)$. Finally, the shared parameters $u_i$ aggregation process is promoted, thereby achieving better performance.}
\end{remark}
\end{theorem}

\section{Experiments}\label{exper}

In this section, we conduct extensive experiments to verify the effectiveness of the proposed DFedAlt and DFedSalt algorithms. Below, we first introduce the experimental setup.

\subsection{Experiment Setup}

% \textbf{Dataset and Data Partition.}
\textbf{Dataset and Data Partition.} We evaluate the performance of our approaches on CIFAR-10, CIFAR-100 \cite{krizhevsky2009learning}, and Tiny-ImageNet \cite{le2015tiny} datasets in the Dirichlet distribution setting and Pathological setting, where CIFAR-10 and CIFAR-100 are two real-life image classification datasets with total 10 and 100 classes. And all detailed experiments on the Tiny-ImageNet dataset are placed in \textbf{Appendix} \ref{exper:tiny} due to the limited space. We partition the training and testing data according to the same Dirichlet distribution Dir($\alpha$) such as $\alpha =0.1$ and $\alpha =0.3$ for each client followed by \cite{hsu2019measuring}. Specifically, the smaller the $\alpha$ is, the more heterogeneous the setting is. Meanwhile, for each client, we sample 2 and 5 classes from a total of 10 classes on CIFAR-10, and 5 and 10 classes from a total of 100 classes on CIFAR-100, respectively \cite{zhang2020personalized}. Where the number of sampling classes is represented as “c” in Table \ref{ta:all_baselines} and the fewer classes each client owns, the more heterogeneous the setting is.

\textbf{Baselines and Backbone.}
We compare the proposed methods with many baselines in both CFL and DFL. For instance, Local is the simplest method where each client only conducts training on their own data without communicating with other clients. And CFL methods include FedAvg \cite{mcmahan2017communication}, FedPer \cite{arivazhagan2019federated} (aka. FedSim \cite{pillutla2022federated}), FedRep \cite{collins2021exploiting} (aka. FedAlt \cite{pillutla2022federated}), FedBABU \cite{oh2021fedbabu}, Fed-RoD\cite{chen2021bridging} and Ditto \cite{li2021ditto}. For DFL mothods, we take DFedAvgM \cite{sun2022decentralized}, Dis-PFL \cite{dai2022dispfl}, DFedSAM \cite{shi2023improving} as our baselines. All methods are evaluated on ResNet-18 \cite{he2016deep} and replace the batch normalization with the group normalization followed by \cite{wu2018group,dai2022dispfl,shi2023improving} to avoid unstable performance. For the partial PFL methods, we set the lower linear layers (close to output) as the personal part responsible for complex pattern recognition, and the rest upper layer (close to input) as the shared part focusing on feature extraction. Note that we compare the personal test accuracy for all methods since our goal is to solve PFL.

\textbf{Implementation Details.}
We keep the same experiment setting for all baselines and perform $500$ communication rounds. The number of  client sizes is 100. The client sampling radio is 0.1 in CFL, while each client communicates with 10 neighbors in PFL accordingly. The batchsize is 128 and the number of local epochs is 5. For DFedAlt and DFedSalt, the local epochs for the shared parameters are 5, while the local epochs of the personal parameters are 1 on Dirichlet dataset and 5 on Pathological dataset. We set SGD\cite{robbins1951stochastic,stich2019local} as the base local optimizer with a learning rate $\eta_v = 0.001$ for the personal and $\eta_u = 0.1$ for shared parameters update with a decay rate of 0.005 and local momentum of $0.9$. Additionally, the weight perturbation ratio in DFedSalt is set to $\rho = 0.7$. We run each experiment 3 times with different random seeds and report the mean accuracy with variance for each method. More details of the baselines can be found in \textbf{Appendix} \ref{exp:baseline}.

\subsection{Performance Evaluation}

\begin{table}[t]
\centering
\scriptsize
\caption{ \small  Test accuracy (\%) on CIFAR-10 \& 100 in both Dirichlet and Pathological distribution settings.}
\label{ta:all_baselines}
\vspace{0.1cm}
\scalebox{0.96}{\begin{tabular}{lcccc|cccc} 
\toprule
\multirow{3}{*}{Algorithm} & \multicolumn{4}{c|}{CIFAR-10}                                                                                          & \multicolumn{4}{c}{CIFAR-100}                                                                                            \\ 
\cmidrule{2-9}
                           & \multicolumn{2}{c}{Dirichlet}                            & \multicolumn{2}{c|}{Pathological}                           & \multicolumn{2}{c}{Dirichlet}       & \multicolumn{2}{c}{Pathological}                                                   \\ 
\cmidrule{2-9}
                           & $\alpha$ = 0.1               & $\alpha$ = 0.3            & c = 2                        & c = 5                        & $\alpha$ = 0.1   & $\alpha$ = 0.3   & c = 5                                                  & c = 10                    \\ 
\midrule
Local                      & $78.96_{\pm.42}$               & $63.20_{\pm.28}$            & $85.16_{\pm.18} $              & $68.56_{\pm.35}$               & $39.38_{\pm.33}$   & $22.59_{\pm.49}$   & $ 71.34_{\pm.46} $                                       & $53.15_{\pm.31} $           \\
FedAvg                     & $84.17_{\pm.28}$               & $80.02_{\pm.20}$            & $84.99_{\pm.11}$               & $81.18_{\pm.27}$               & $57.35_{\pm.03}$   & $55.12_{\pm.06} $  & $ 69.29_{\pm.43} $                                       & $66.10_{\pm.48} $           \\
FedPer                     & $88.57_{\pm.09}$               & $84.06_{\pm.29}$            & $90.94_{\pm.24}$               & $86.97_{\pm.35}$               & $54.23_{\pm.14}$   & $34.07_{\pm.76}$   & $ 78.48_{\pm.93 }$                                       & $ 70.38_{\pm.02} $          \\
FedRep                     & $88.78_{\pm.40}$               & $84.50_{\pm.05}$            & $91.09_{\pm.12} $              & $ 86.22_{\pm.51}$              & $ 44.02_{\pm.98 }$ & $26.88_{\pm.49}$   & $78.77_{\pm.19}$                                         & $68.15_{\pm.43 }$           \\
FedBABU                    & $87.79_{\pm.53}$               & $83.26_{\pm.09}$            & $91.32_{\pm.15} $              & $ 84.90_{\pm.24} $             & $60.23_{\pm.07} $  & $52.37_{\pm.82}$   & $ 77.50_{\pm.33} $                                       & $69.81_{\pm.12} $           \\
Fed-RoD                    & $89.15_{\pm.12}$               & $85.68_{\pm.08}$            & $90.10_{\pm.04} $              & $87.81_{\pm.45}$                        & $ 65.79_{\pm.05} $ & $58.54_{\pm.69} $  & $ 80.50_{\pm.45} $                                       & $ 73.59_{\pm.15} $          \\
Ditto                      & $80.22_{\pm.10}$               & $73.51_{\pm.04}$            & $84.96_{\pm.40} $              & $ 75.59_{\pm.32} $             & $48.85_{\pm.54 }$  & $48.65_{\pm.50} $  & $69.48_{\pm.45} $                                        & $ 60.77_{\pm.30} $          \\ 
\midrule
DFedAvgM                   & $87.39_{\pm.13}$               & $ 82.60_{\pm.18} $          & $90.72_{\pm.08 }$              & $84.69_{\pm.25} $ & $ 59.76_{\pm.69}$  & $ 54.98_{\pm.48} $ & $76.70_{\pm.59} $                                        & $ 71.08_{\pm.52} $          \\
Dis-PFL                     & $87.77_{\pm.46 }$ & $82.71_{\pm.28}$            & $88.19_{\pm.47}$               & $82.29_{\pm.61}$             & $56.06_{\pm.20}$   & $46.65_{\pm.18} $   & $71.79_{\pm.42} $                                        & $65.35_{\pm.10 }$           \\
DFedSAM                    & $84.96_{\pm.30} $              & $77.36_{\pm.11}$            & $90.14_{\pm.22}$ & $83.05_{\pm.40}$             & $58.21_{\pm.53 }$  & $47.80_{\pm.49} $  & $74.25_{\pm.17} $                                        & $67.34_{\pm.43} $           \\ 
\midrule
DFedAlt                    & $88.85_{\pm.21}$               & $86.50_{\pm.05} $           & $91.26_{\pm.23} $              & $86.85_{\pm.37} $              &$66.26_{\pm.25}$                  & $57.66_{\pm.42}$                 & $78.78_{\pm.41}$                           & $72.19_{\pm.21} $           \\
DFedSalt                   & $\textbf{91.08}_{\pm.34}$      & $ \textbf{87.67}_{\pm.22} $ & $ \textbf{92.20}_{\pm.14} $    & $\textbf{88.34}_{\pm.31} $    &  $\textbf{67.03}_{\pm.36}$                 & $\textbf{ 58.73}_{\pm.19} $                 & $\textbf{80.82}_{\pm.33}$ & $\textbf{74.50}_{\pm.35} $  \\
\bottomrule
\end{tabular}}
\vspace{-0.3cm}
\end{table}

\begin{figure}[t]
    \centering
    % \vspace{-0.4cm}
    \begin{subfigure}{1\linewidth}
		\centering		\includegraphics[width=1.0\linewidth]{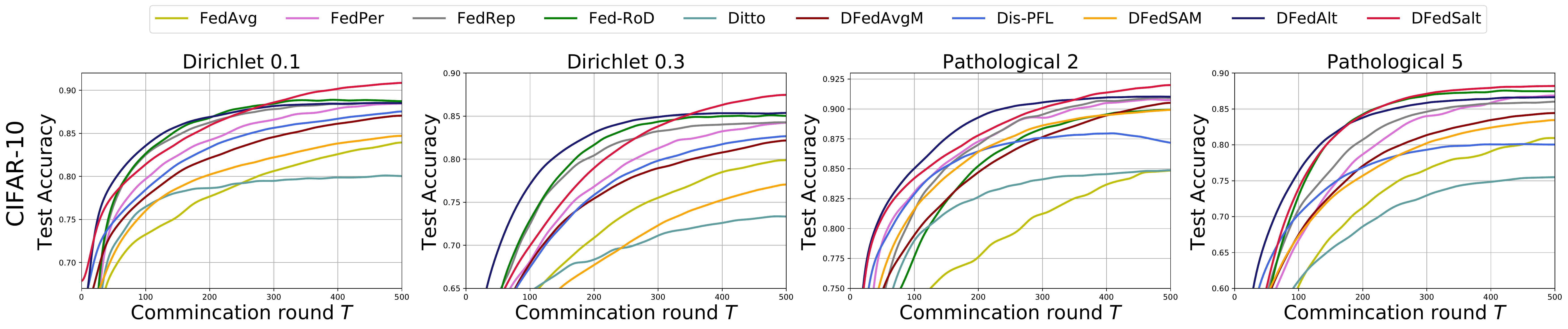}
		%\caption{\small CIFAR-10}
	    \label{CIFAR-10}
     \vspace{-0.3cm}
	\end{subfigure}
         \begin{subfigure}{1\linewidth}
		\centering		\includegraphics[width=1.0\linewidth]{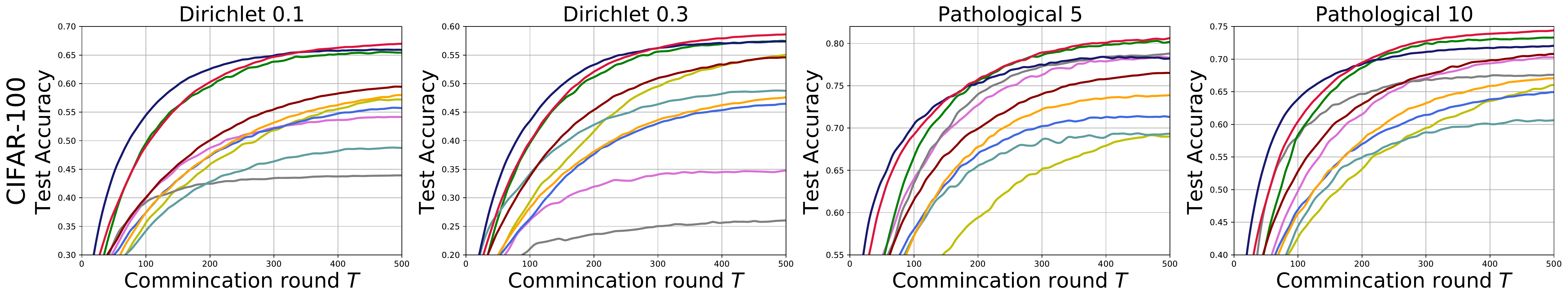}
		%\caption{\small CIFAR-100}
	    \label{CIFAR-100}
        \vspace{-0.3cm}
	\end{subfigure}
        \caption{\small Test accuracy on CIFAR-10 (first line) and CIFAR-100 (second line) with heterogenous data partitions.}
	\label{fig:baseline}
     \vspace{-0.5cm}
\end{figure}

\textbf{Comparison with the baselines.} As shown in Table \ref{ta:all_baselines} and Figure \ref{fig:baseline}, the proposed DFedAlt and DFedSalt outperform other baselines with the best stability and perform well in severe data heterogeneity scenarios. It significantly proves that decentralized training is more suitable for PFL than centralized training. Specifically, on the CIFAR-10 dataset, DFedAlt and DFedSalt achieve 87.67\% and 86.50\% on the Directlet-0.3 setups, 0.82\% and 1.99\% ahead of the best comparing CFL method Fed-RoD. On the CIFAR-100 dataset, DFedSalt achieves at least 0.32\% and 0.91\% improvement from the other baselines on the Pathological-5 and Pathological-10 settings. The effect of the hyper network in Fed-RoD is remarkable but partial model personalization in the decentralized scenario can obtain greater gains than it. DFedAlt and DFedSalt focus on local optimization and absorb the feature extraction capabilities learned by other users on their own data. So they maintain the classified head more adapted to the local data for each client with a stronger feature extractor.

\textbf{Discussion on the heterogeneous setting.} We discuss two different data heterogeneity of Dirichlet distribution and Pathological distribution in Table \ref{ta:all_baselines}, and we further prove the effectiveness and robustness of the proposed methods. In Dirichlet distribution, since the local training is hard to cater for all classes inside clients, the accuracy decreases with the level of heterogeneity decreasing. On CIFAR-10, when the heterogeneity increases from 0.1 to 0.3, Fed-RoD drops from 89.15\% to 85.68\%, while DFedSalt drops about 3.41\% to 87.67\%, meaning its strong adaptability and stability for several heterogeneous settings. Pathological distribution defines limited classes for each client which is a higher level of heterogeneity. In detail, DFedSalt is 0.88\% ahead of the best compared CFL method on CIFAR-10 with only 2 categories per client and 0.91\% ahead on CIFAR-100 dataset with only 10 categories per client. The comparisons confirm that the proposed methods could achieve better performance in the strong heterogeneity.

\begin{wraptable}[16]{r}{0.45\linewidth}
\scriptsize
\vspace{-0.4cm}
\centering
\caption{ \small The required communication rounds when achieving the target accuracy (\%).}
\vspace{-0.2cm}
\label{ta:convergency speed}
\begin{tabular}{lcc|cc} 
\toprule
\multirow{3}{*}{Algorithm} & \multicolumn{2}{c|}{CIFAR-10}   & \multicolumn{2}{c}{CIFAR-100}    \\ 
\cmidrule{2-5}
                           & Dir-0.3        & Pat-2          & Dir-0.3        & Pat-10          \\ 
\cmidrule{2-5}
                           & acc@80         & acc@90         & acc@45         & acc@65          \\ 
\midrule
FedAvg                     & -              & -              & 234            & 456             \\
FedPer                     & 262            & 343            & -              & 246             \\
FedRep                     & 189            & 322            & -              & 225             \\
FedBABU                    & 270            & 312            & 261            & 314             \\
Fed-RoD                    & 170            & 462            & 133  & 148             \\
Ditto                      & -              & -              & 279            & -               \\ 
\midrule
DFedAvgM                    & 354            & 439            & 192            & 230             \\
Dis-PFL                    & 307            & -              & 368            & 492             \\
DFedSAM                    & -              & 465            & 367            & 344             \\ 
\midrule
DFedAlt                    & \textbf{ 131 } & \textbf{ 224 } & \textbf{111}          & \textbf{ 113 }  \\
DFedSalt                   & 160            & 280            & 131               & 139             \\
\bottomrule
\end{tabular}
\end{wraptable}

\begin{wrapfigure}[10]{r}{0.5\linewidth}
    \centering
     \vspace{-0.4cm}
    \begin{subfigure}{1.0\linewidth}
		\centering
          \includegraphics[width=1.0\textwidth]{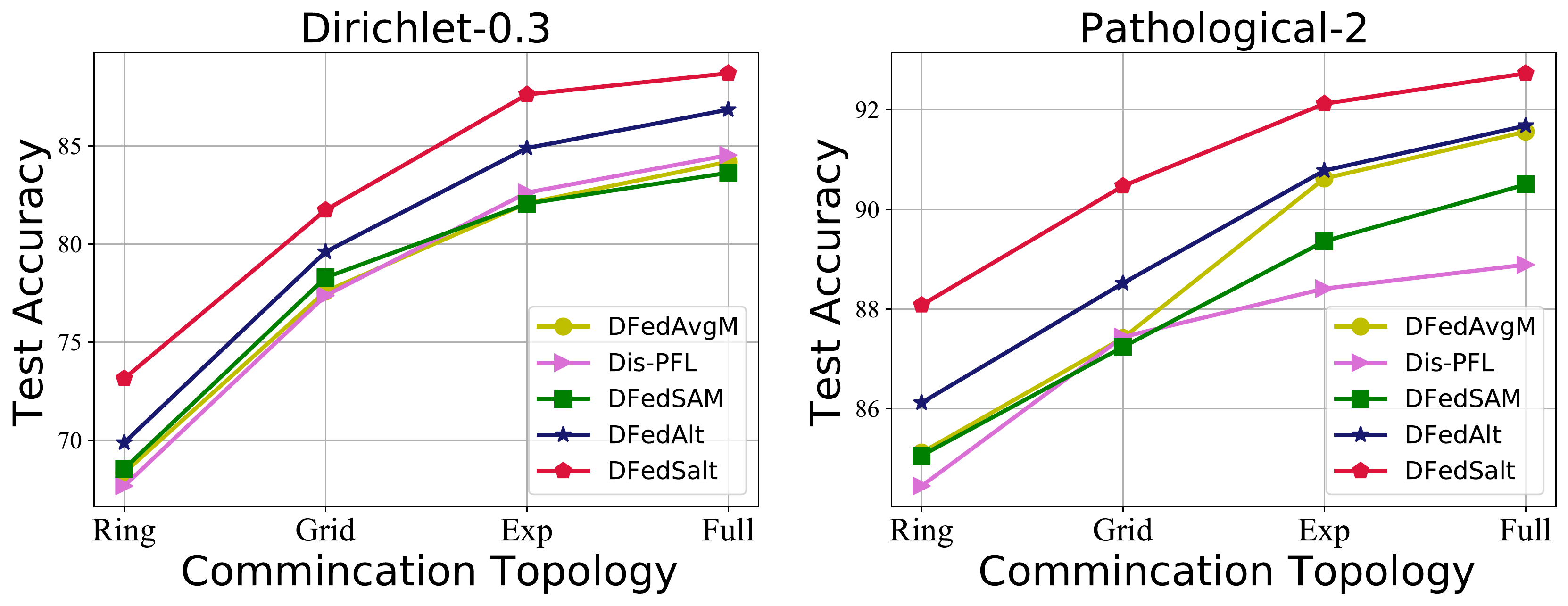}
	\end{subfigure}
\caption{\small Personal test accuracy (\%) in various network topologies for the DFL methods on CIFAR-10.}
\label{fig:top}
\end{wrapfigure}

\textbf{Convergence speed.} We illustrate the convergence speed for all baselines via the learning curves under different settings in Figure \ref{fig:baseline} and collect the communication rounds for each method to reach a target accuracy (acc@) in Table \ref{ta:convergency speed}. The results show that DFedAlt achieves the fastest convergence speed among the comparison methods, which benefits from the decentralized training mode and alternate update a lot. In comparison with the indirect interaction methods in CFL, which pass information through a central node, the direct information interaction in DFL can speed up the convergence rate for personalized problems. Also, the difference between DFedAlt and DFedAvgM indicates that the convergence speed of alternate updating is faster than uniform updating. Besides, Figure \ref{fig:top} also reveals that the fully-connected case may outperform other communication topologies in convergence speed. This can be attributed to the increased communication information per round, which has also been verified in \cite{koloskova2019decentralized,dai2022dispfl}. Notably, we target the setting where the busiest node’s communication bandwidth is restricted for fairness when compared with the CFL methods.

\textbf{Discussion on communication topologies.} In practice, the clients are often connected with pre-given topologies and have different computing and communication capabilities, named heterogeneous clients. For decentralized methods, comparing the performance of various communication topologies will help us evaluate the performance of the methods with heterogeneous clients. Figure  \ref{fig:top} shows the performances of each decentralized method in various communication topologies on the CIFAR-10 dataset. The sparse degree of the communication topology from sparse to compact are Ring, Grid, Exp (aka. exponential), and Full (aka. full-connected), and the performance of all the methods improves correspondingly with the communication topology compacter. Besides, we find that the proposed DFedAlt and DFedSalt perform extremely well in various communication topologies. Specifically, in the Dirichlet setting, DFedAlt and DFedSalt outperform other baselines from 1.31\%-5.00\%. In the pathological setting, the accuracy gap between the Ring and Full topology of DFedAvg, DFedAlt, and DFedSalt is 6.44\%, 5.56\%, and 4.65\%, respectively. This indicates that the DFedAlt and DFedSalt are more robust and suitable for the DFL setting. 

\subsection{Ablation Study}

\begin{wraptable}{r}{0.5\linewidth}
\centering
\vspace{-0.4cm}
\scriptsize
\caption{ \small  Test accuracy (\%) of different model parts with the SAM optimizer.}
\label{ta:samwhere}
\vspace{-0.2cm}
\begin{tabular}{l|cc|c!{\vrule width \lightrulewidth}l} 
\toprule
Algorithm  & Body & Head & Dirichlet~     & Pathological                        \\ 
\midrule
DFedAlt    &      &      & 86.50          & \multicolumn{1}{c}{91.26}           \\
DFedSalt-U  & \checkmark     &      & \textbf{87.67} & \multicolumn{1}{c}{92.20}  \\
DFedSalt-V  &      & \checkmark      & 86.46        & \multicolumn{1}{c}{91.50}                               \\
DFedSalt-UV & \checkmark      &\checkmark       & 87.43            & \multicolumn{1}{c}{\textbf{{92.58}}}                                \\
\bottomrule
\end{tabular}
\vspace{-0.2cm}
\end{wraptable}

\textbf{Integrating SAM into the shared model $u_i$ or personal model $v_i$ or whole model $(u_i, v_i)$.} We investigate the effect of adding the SAM optimizer to different parts with different data heterogeneity on the CIFAR-10 dataset. From Table \ref{ta:samwhere}, DfedSalt-U (SAM only for the shared model, dubbed as “body”) achieves the best in Dirichlet setting and DFedSalt-UV (adding SAM to both shared and personal parts) achieves the best in Pathological setting. From the difference between DFedAlt, DFedSalt-U and DFedSalt-UV, we observe that the SAM optimizer can uniformly reduce the inconsistency of the feature extractor among clients and improve the feature extraction ability of the shared parts. Besides, the comparison from DFedAlt, DFedSalt-V (SAM only for the personal model, dubbed as “head”) and DFedSalt-UV illustrates that the benefits of adding SAM to the personal model may be sensitive to the data distribution and hyperparameter setup. Thus, in the main experiments, we set DFedSalt-U as our default algorithm and denote it as DFedSalt.

%\textbf{Number of Local epochs for the personal model.} We study the number of local epochs for the personal model in Figure\ref{}. The most suitable local epochs for the personal model on Dir and Pathological dataset are different.

%\begin{wrapfigure}[12]{r}{0.5\linewidth}
%    \centering
%     \vspace{-0.3cm}
%    \begin{subfigure}{1.0\linewidth}
%		\centering
%          \includegraphics[width=1.0\textwidth]{images/top.pdf}
%	\end{subfigure}
%\caption{\small Number of local epochs——the experiments is running.}
%\label{fig:local epochs}
%\end{wrapfigure}

\begin{wrapfigure}[10]{r}{0.55\linewidth}
    \centering
     \vspace{-0.5cm}
    \begin{subfigure}{1.0\linewidth}
		\centering
          \includegraphics[width=1.0\textwidth]{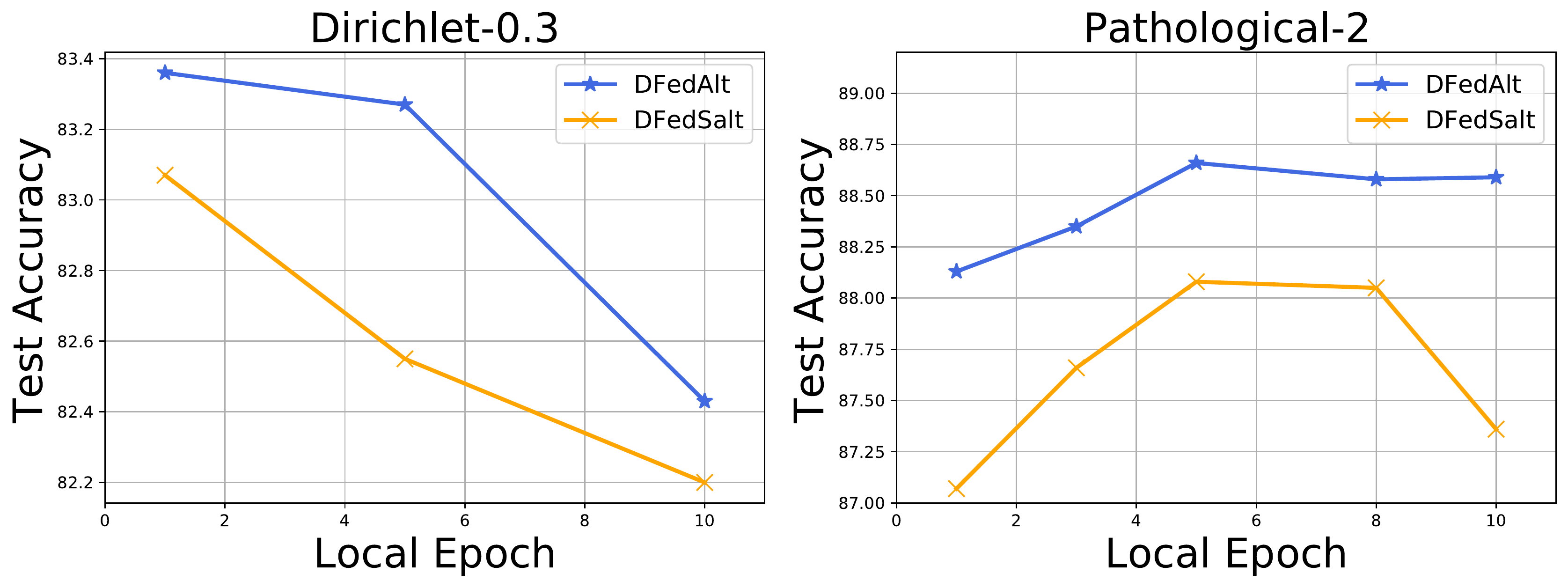}
	\end{subfigure}
\caption{\small Effect of the local epochs for the personal parameters $v_i$ in client $i$.}
\label{fig:abla epoch}
   \vspace{-0.15cm}
\end{wrapfigure}

\textbf{Effectiveness of local epochs.} In Figure \ref{fig:abla epoch}, we illustrate the effect of local epochs for the personal parameters in different heterogeneity scenarios on the CIFAR-10 dataset after 200 communication rounds. For the Dirichlet scenarios, with fixed local epochs of 5 for the shared parameters, more local epochs (i.e., larger $K_v$) for the personal parameters will damage the performance. That means fewer local epochs for the personal improve the shared part more and the personal part with less relative variance per user also fits well on local data. While in the Pathological scenarios, a more heterogeneous distribution for each client, the local epochs for the personal parameters must be a trade-off to improve the extraction ability of the shared part and adapt the personal part to the local data.

\begin{wrapfigure}[8]{r}{0.43\linewidth}
    \centering
     \vspace{-0.7cm}
    \begin{subfigure}{1.0\linewidth}
		\centering          \includegraphics[width=0.85\textwidth]{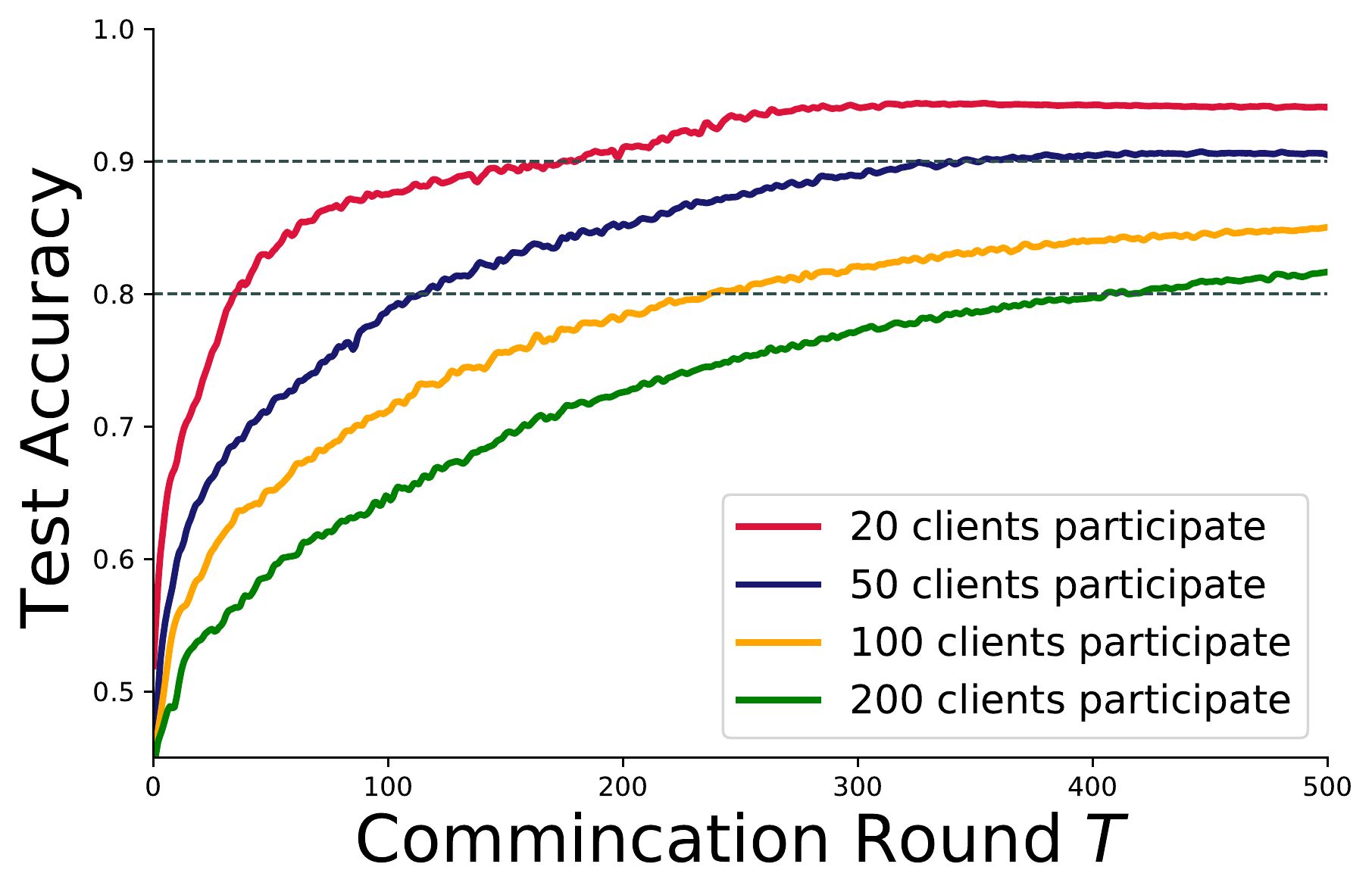}
	\end{subfigure}
 \vspace{-0.6cm}
\caption{\small Effect of the participated clients size.}
\label{fig:abla client number}
\end{wrapfigure}

\textbf{Number of participated clients.} We compare the performance between different numbers of client participation of \{20, 50, 100, 200\} on CIFAR-10 with Dirichlet $\alpha=0.3$ in Figure \ref{fig:abla client number}. It is clearly seen that the test performance will get a great margin with the participation of clients decreasing, which means that the more training data each client owns, the better performance it will achieve.

\section{Conclusion}
In this paper, we propose two novel methods --- DFedAlt and DFedSalt for PFL, which improve the representation ability via adopting decentralized partial model personalization to seek out suitable personalization in FL. It efficiently personalizes the “right” components in the deep modern models and alternatively updates the shared parameters and personal parameters in a peer-to-peer manner. For theoretical findings, we present the convergence rate in the stochastic non-convex setting for DFedAlt and DFedSalt. Empirical results also verify the superiority of our approaches. 

\textbf{Limitations.} In the current version, we mainly focus on the  optimizer problem in personalized federated learning and have not explored further research and analysis on the generalization ability of the proposed methods. We will perform continuous research in this aspect in future work.

\textbf{Broader Impacts.} In this paper, we analyze and explore the main problems existing in personalized federated learning: single point of failure, communication burden, and poor representation ability. And we adopt the decentralized partial model personalization technique to alleviate the above-mentioned problems, dubbed as DFedAlt and DFedSalt. Meanwhile, we give theoretical convergence analysis and extensive experiments, these attempts indicate the effectiveness of our approaches.

{
% \clearpage
% \small
\bibliographystyle{plainnat} %{ieee_fullname}
\bibliography{ref}
}

\clearpage
\begin{center}
 \rule{5.6in}{0.7pt}\\ % 4.0
\vspace{0.2cm}
 {\Large\bf Supplementary Material for \\`` Towards More Suitable Personalization in Federated Learning via Decentralized Partial Model Training ''}
 \rule{5.6in}{0.7pt}
\end{center}
\appendix

In this part, we provide the supplementary materials including more introduction to the related works, experimental details and results, and the proof of the main theorem.
\begin{itemize}
    \item \textbf{Appendix} \ref{ap:related_works}: More details in the related works.
    \item \textbf{Appendix} \ref{exp:baseline}: More details in the experiments.
    \item \textbf{Appendix} \ref{ap:proof}: Proof of the theoretical analysis.
\end{itemize}
\section{More Details in the Related Works}\label{ap:related_works}

\textbf{Decentralized/Distributed Training.}
By combining SGD and gossip, early work achieved decentralized training and convergence of the model  in \cite{blot2016gossip}. D-PSGD \cite{lian2017can} is the classic decentralized parallel SGD method. FastMix \cite{ye2020decentralized} investigates the advantage of increasing the frequency of local communications within a network topology, in which the optimal computational complexity and near-optimal communication complexity are established. DeEPCA \cite{ye2021deepca} integrates FastMix into a decentralized PCA algorithm to accelerate the training process. DeLi-CoCo \cite{Hashemi2022On} performs multiple compression gossip steps in each iteration for fast convergence with arbitrary communication compression. Network-DANE \cite{li2020communication} uses multiple gossip steps and generalizes DANE to decentralized scenarios.
The work in \cite{lin2021quasi} modifies the momentum term of decentralized SGD (DSGD) to be adaptive to heterogeneous data, while the work in \cite{hsieh2020non} replaces batch norm with layer norm. \cite{li2022learning} dynamically updates the mixing weights based on meta-learning and learns a sparse topology to reduce communication costs. The work in \cite{zhu2022topology} provides the topology-aware generalization analysis for DSGD, they explore the impact of various communication topologies on the generalizability.

\textbf{Sharpness Aware Minimization (SAM).} 
SAM \cite{foret2021sharpnessaware} is an effective optimizer for training deep learning (DL) models, which leverages the flatness geometry of the loss landscape to improve model generalization ability. Recently, the work in \cite{Andriushchenko2022Towards} studies the properties of SAM and provides convergence results of SAM for non-convex objectives. As a powerful optimizer, SAM and its variants have been applied to various DL tasks \cite{Zhao2022Penalizing,kwon2021asam,du2021efficient,liu2022towards,Abbas2022Sharp-MAML,mi2022make,zhong2022improving,huangrobust} and FL tasks \cite{Qu2022Generalized,Caldarola2022Improving,sunfedspeed,sun2023adasam,shi2023improving,shi2023towards,shi2023make}. For instance, the works in \cite{Qu2022Generalized}, \cite{sunfedspeed}, and \cite{Caldarola2022Improving} integrate SAM to improve the generalization, and thus mitigate the distribution shift and achieve a new SOTA performance for FL.

\section{More Details in the Experiment}\label{exp:baseline}
In this section, we provide more details of our experiments and more extensive experimental results to compare the performance of the proposed DFedAlt and DFedSalt against other baselines.

\subsection{Datasets and Data Partition}

\begin{table}[ht]
\centering
\small
\caption{ \small  The details on the CIFAR-10 and CIFAR-100 datasets.}
\label{ta:all_data}
\begin{tabular}{ccccc} 
\toprule
Dataset       & Training Data & Test Data & Class & Size     \\ 
\midrule
CIFAR-10      & 50,000        & 10,000    & 10    & 3×32×32  \\
CIFAR-100     & 50,000        & 10,000    & 100   & 3×32×32  \\
Tiny-ImageNet & 100,000       & 10,000    & 200   & 3×64×64        \\
\bottomrule
\end{tabular}
\end{table}

CIFAR-10/100 and Tiny-ImageNet are three basic datasets in the computer version study. As shown in Table \ref{ta:all_data}, they are all colorful images with different classes and different resolutions. We use two non-IID partition methods to split the training data in our implementation. One is based on Dirichlet distribution on the label ratios to ensure data heterogeneity among clients, where a smaller $\alpha$ means higher heterogeneity. Another assigns each client a limited number of categories, called Pathological distribution, where fewer categories mean higher heterogeneity. The distribution of the test datasets is the same as in training datasets. We run 500 communication rounds for CIFAR-10, CIFAR-100, and 300 rounds for Tiny-ImageNet.
% \subsection{Communication Topology}
\subsection{More Details about Baselines}
\textbf{Local} is the simplest method for personalized learning. It only trains the personalized model on the local data and does not communicate with other clients. For the fair competition, we train 5 epochs locally in each round. 

\textbf{FedAvg}\cite{mcmahan2017communication} is the most commonly discussed method in FL. It selects some clients to perform local training on each dataset and then aggregates the trained local models to update the global model. Actually, the local model in FedAvg is also the comparable personalized model for each client. 

\textbf{FedPer}\cite{arivazhagan2019federated} proposes a base + personalized layer approach for PFL to combat the ill effects of statistical heterogeneity. We set the linear layer as the personalized layer and the rest model as the base layer.  It follows FedAvg’s training paradigm but only passes the base layer to the server and keeps the personalized layer locally.

\textbf{FedRep}\cite{collins2021exploiting} also proposes a body(base layer) + head(personalized layer) framework like FedPer, but it fixes one part when updating the other. We follow the official implementation\footnote{\url{https://github.com/lgcollins/FedRep}} to train the head for 10 epochs with the body fixed, and then train the body for 5 epochs with the head fixed. 

\textbf{FedBABU}\cite{oh2021fedbabu} is also a model split method that achieves good personalization via fine-tuning from a good shared representation base layer. Different from FedPer and FedRep, FedBABU only updates the base layer with the personalized layer fixed and finally fine-tunes the whole model. Following the official implementation\footnote{\url{https://github.com/jhoon-oh/FedBABU}}, it fine-tunes 5 times in our experiments.

\textbf{Fed-RoD}\cite{chen2021bridging} explicitly decouples a model’s dual duties with two prediction tasks---generic optimization and personalized optimization and utilizes a hyper network to connect the generic model and the personalized model. Each client first updates the generic model with balanced risk minimization then updates the personalized model with empirical risk minimization. 

\textbf{Ditto}\cite{li2021ditto} achieves personalization via a trade-off between the global model and local objectives. It totally trains two models on the local datasets, one for the global model (similarly aggregated as in FedAvg) with its local empirical risk, and one for the personal model (kept locally) with both empirical risk and the proximal term towards the global model. We set the regularization parameters $\lambda$ as 0.75. 

\textbf{DFedAvgM}\cite{sun2022decentralized} is the decentralized FedAvg with momentum, in which clients only connect with their neighbors by an undirected graph. For each client, it first initials the local model with the received models then updates it on the local datasets with a local stochastic gradient. 

\textbf{DFedSAM}\cite{shi2023improving} leverages gradient perturbation to generate local flat models via Sharpness Aware Minimization (SAM). The communication framework between neighbors is the same as DFedAvgM, but the local update is performed by the SAM optimizer. We set the perturbation radius $\rho=0.01$ in our experiments followed by \cite{shi2023improving}. 

\textbf{Dis-PFL}\cite{dai2022dispfl} employs personalized sparse masks to customize sparse local models in the PFL setting. Each client first initials the local model with the personalized sparse masks and updates it with empirical risk. Then filter out the parameter weights that have little influence on the gradient through cosine annealing pruning to obtain a new mask. Following the official implementation\footnote{\url{https://github.com/rong-dai/DisPFL}}, the sparsity of the local model is set to 0.5 for all clients.
  
\subsection{More Experiments Results on Tiny Imagenet}\label{exper:tiny}

\begin{figure}[ht]
\centering
\includegraphics[width=1\textwidth]{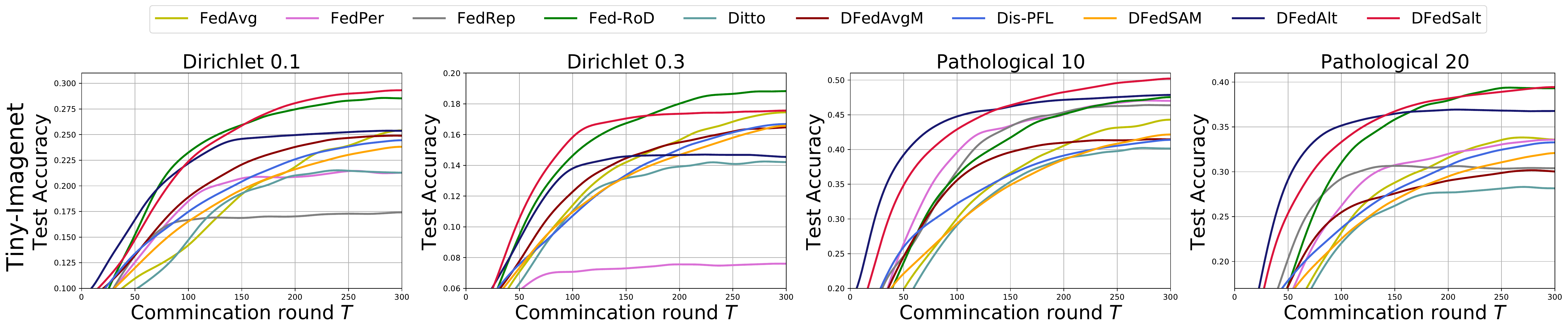}
\centering
\vspace{-0.35cm}
\caption{\small Test accuracy on Tiny-ImageNet with heterogenous data partitions. }
\label{img_tinybaseline}
\vspace{-0.35cm}
\end{figure}

\begin{table}[h]
\centering
\scriptsize
\caption{ \small  Test accuracy (\%) on Tiny-ImageNet in both Dirichlet and Pathological distribution settings on Tiny-ImageNet.}
\label{ap:ta_baselines}
\begin{tabular}{lcccc} 
\toprule
\multirow{3}{*}{Algorithm} & \multicolumn{4}{c}{Tiny-ImageNet}                                                                            \\ 
\cmidrule{2-5}
                           & \multicolumn{2}{c}{Dirichlet}                       & \multicolumn{2}{c}{Pathological}                      \\ 
\cmidrule{2-5}
                           & $\alpha$ = 0.1          & $\alpha$ = 0.3            & c = 10                    & c = 20                     \\ 
\midrule
Local                      & $12.13_{\pm.13}$          & $5.42_{\pm.21} $        & $28.49_{\pm.16} $           & $16.72_{\pm.34} $            \\
FedAvg                     & $25.55_{\pm.02}$          & $17.58_{\pm.25}$            & $44.56_{\pm.39}$            & $ 34.10_{\pm.59}$             \\
FedPer                     & $21.64_{\pm.72}$          & $7.71_{\pm.08} $            & $47.35_{\pm.03}$            & $ 33.68_{\pm.33}$             \\
FedRep                     & $17.54_{\pm.79}$          & $5.78_{\pm.05}$            & $46.76_{\pm.73} $           & $ 31.15_{\pm.54}$             \\
FedBABU                    & $27.40_{\pm.08}$          & $\textbf{19.73}_{\pm.06}$            & $46.53_{\pm.20} $           & $ 38.68_{\pm.31} $           \\
Fed-RoD                    & $29.03_{\pm.55}$          & $19.25_{\pm.45} $           & $48.01_{\pm.40} $           & $ 39.28_{\pm.58} $           \\
Ditto                      & $21.71_{\pm.66}$          & $14.47_{\pm.14}$            & $40.65_{\pm.15} $           & $ 28.74_{\pm.38} $           \\ 
\midrule
DFedAvgM                   & $25.29_{\pm.26}$          & $17.07_{\pm.17} $           & $42.80_{\pm.43} $           & $30.58_{\pm.51}$         \\
Dis-PFL                     & $24.71_{\pm.18}$          & $16.94_{\pm.36}$            & $41.93_{\pm.12}$            & $33.57_{\pm.62}$        \\
DFedSAM                    & $24.18_{\pm.32}$          & $16.92_{\pm.19}$            & $42.87_{\pm.31}$            &  $32.61_{\pm.14}$           \\ 
\midrule
DFedAlt                    & $25.71_{\pm.20}$          & $14.94_{\pm.44}$          & $49.16_{\pm.19} $           & $37.25_{\pm.27}  $            \\
DFedSalt                   & $\textbf{29.70}_{\pm.47}$ & $ 17.81_{\pm.35} $ &$\textbf{50.79}_{\pm.28}$ & $\textbf{39.44}_{\pm.40}$ \\
\bottomrule
\end{tabular}
\end{table}

\textbf{Comparison with the baselines.} In Table \ref{ap:ta_baselines} and Figure \ref{img_tinybaseline}, we compare DFedAlt and DFedSalt with other baselines on the Tiny-ImageNet with different data distributions. The comparison shows that the proposed methods have a competitive performance, especially under higher heterogeneity, e.g. for Dirichlet-0.1 and Pathological-10. Specifically in the Dirichlet-0.1 setting, DFedSalt achieves 29.70\%, at least 0.67\% improvement from the CFL methods, while DFedSalt and DFedAlt are 1.15\% and 2.78\%  ahead of the other baselines in Pathological-10 setting. The original intention of our design is to build a great personalized model by focusing on local training and exchanging the feature extraction capabilities with neighbors via decentralized partial model training. So when the heterogeneity increases, our algorithms have a significant improvement.

\begin{table}[ht]
\centering
\scriptsize
\caption{ \small The required communication rounds when achieving the target accuracy (\%) on Tiny-ImageNet.}
\label{ap:ta_convergency speed}
\begin{tabular}{lcc|cc|cc|cc} 
\toprule
\multirow{3}{*}{Algorithm} & \multicolumn{8}{c}{Tiny-ImageNet}                                                                                     \\ 
\cmidrule{2-9}
                           & \multicolumn{2}{c|}{Dirichlet-0.1} & \multicolumn{2}{c|}{Dirichlet-0.3} & \multicolumn{2}{c|}{Pathological-10} & \multicolumn{2}{c}{Pathological-20}  \\ 
\cmidrule{2-9}
                           & acc@20      & speedup       & acc@14 & speedup             & acc@40      & speedup      & acc@30      & speedup       \\ 
\midrule
FedAvg                     & 160         & 1.11 ×         & 144    & 1.47 ×              & 192         & 1.36 ×        & 172         & 1.50 ×         \\
FedPer                     & 123         & 1.45 ×         & -      & -                   & 103         & 2.53 ×        & 134         & 1.93 ×         \\
FedRep                     & -           & -             & -      & -                   & 116         & 2.25 ×        & 117         & 2.21 ×         \\
FedBABU                    & 156         & 1.14 ×         & 174    & 1.22 ×               & 178         & 1.47 ×        & 181         & 1.43 ×         \\
Fed-RoD                    & \textbf{72} & \textbf{2.47 ×}        & 92     & 2.30 ×                & 132         & 1.98 ×        & 95          & 2.72 ×         \\
Ditto                      & 178         & 1.00 ×            & 212    & 1.00 ×                  & 261         & 1.00 ×           & -           & -             \\ 
\midrule
DFedAvgM                   & 115         & 1.55 ×         & 136    & 1.56 ×            & 160         & 1.63 ×        & 258       & 1.00 ×            \\
Dis-PFL                    & 143         & 1.24 ×         & 166    & 1.28 ×            & 227         & 1.15 ×        & 188       & 1.37 ×               \\
DFedSAM                    & 158         & 1.13 ×         & 174    & 1.22 ×            & 229         & 1.14 ×        & 214       & 1.21 ×              \\ 
\midrule
DFedAlt                    & 74          & 2.41 ×         & 108    & 1.96 ×               & \textbf{54} & \textbf{4.83 ×}      & \textbf{53} & \textbf{4.87 × }        \\
DFedSalt                   & 82          & 2.17 ×         & \textbf{78}       &  \textbf{2.72 × }            & 70          & 3.73 ×        & 73          & 3.53 ×         \\
\bottomrule
\end{tabular}
\end{table}

\textbf{Convergence speed.} We show the convergence speed of DFedAlt and DFedSalt in Table \ref{ap:ta_convergency speed} by reporting the number of rounds required to achieve the target personalized accuracy (acc@) on Tiny-ImageNet. For each setting, we set the algorithm that takes the most rounds to reach the target accuracy as “1.00×”, and find that the proposed DFedAlt and DFedSalt achieve the fastest convergence speed on average (3.51× and 3.04× on average) among the SOTA PFL algorithms. Local training in PFL consistently pursues empirical risk minimization on the local datasets, which can efficiently train the personalized model fitting the local distribution. Also, the alternate updating mode will bring a comparable gain to the convergence speed from the difference between DFedAlt and DFedAvgM. Thus, our methods can efficiently train the personalized model, especially on the higher heterogeneity.

% \subsubsection{Loss landscape}
% To visualize the sharpness of the flat minima and observe the improvement of the feature extraction capability of SAM, we show the loss landscape and surface contour following \cite{li2018visualizing} in Figure \ref{}. The models are trained on CIFAR-10 with the Dirichlet-0.3 setup. It is clear see that DFedSalt has not only a lower loss value but also a smoother loss landscape, which indicates that the SAM optimization can be effectively applied to PFL through its excellent generalization performance in feature learning.

\section{Proof of Theoretical Analysis}\label{ap:proof}

\subsection{Preliminary Lemmas}
\begin{lemma}[Lemma 4, \cite{lian2017can}]
\label{mi}
For any $t\in \mathbb{Z}^+$, the mixing matrix ${\bf W}\in\ \mathbb{R}^m$ satisfies
$\|{\bf W}^t-{\bf P}\|_{\emph{op}}\leq \lambda^t,$
where $\lambda:=\max\{|\lambda_2|,|\lambda_m(W)|\}$ and for a matrix ${\bf A}$, we denote its spectral norm as $\|{\bf A}\|_{\emph{op}}$. Furthermore, ${\bf 1}:=[1, 1, \ldots, 1 ]^{\top}\in \mathbb{R}^m$ and
\begin{equation*}
    {\bf P}:=\frac{\mathbf{1}\mathbf{1}^{\top}}{m}\in \mathbb{R}^{m\times m}.
\end{equation*}
\end{lemma}

\begin{lemma}[Lemma 23, \cite{pillutla2022federated}]
\label{le:v} 
    Consider $F$ which is $L$-smooth and fix a $v^0 \in \mathbb{R}^d$. 
	Define the sequence $(v^{k})$ of iterates produced by stochastic gradient descent with a fixed learning rate $\eta_v \leq 1/(2K_vL_v)$
	starting from $v^{0}$, we have the bound
	\[
		\E \| v^{K_v-1} - v^{0} \|^2 \le 16 \eta_v^2 K_v^2 \E \|\nabla  F(v^{0})\|^2 + 8 \eta_v^2 K_v^2 \sigma_v^2 \,.
        \]
\end{lemma}
\begin{lemma}[Local update for shared model $u_i$ in DFedAlt]
\label{le:u_i_t_k_alt}
Assume that assumptions 1-3 hold, for all clients $i \in \{1,2,...,m\}$ and local iteration steps $k \in \{0,1,..., K_u-1\}$, we can get
\begin{equation}
    \frac{1}{m}\sum_{i=1}^m \E \bigl\| u_i^{t,k} - u_i^t\bigr \|^2 \leq 18\eta_u^2K_u^2\Big(\sigma_u^2+\delta^2+\frac{1}{m}\sum_{i=1}^m\E \bigl\| \nabla_u F(u_i^t,V^{t+1}) \bigr \|^2\Big). \nonumber
\end{equation}
\begin{proof}
\begin{equation}
    \begin{split}
        \frac{1}{m}\sum_{i=1}^m \E \bigl\| u_i^{t,k+1} - u_i^t\bigr \|^2 
        & \leq \frac{1}{m}\sum_{i=1}^m  \E \Big\| u_i^{t,k} - \eta_u \nabla_u F_i(u_i^{t,k}, v_i^{t+1};\xi_i) - u_i^t \Big\|^2\\
        & \leq \frac{1}{m}\sum_{i=1}^m  \E \Big\| u_i^{t,k} - u_i^{t} - \eta_u \Big( \nabla_u F_i(u_i^{t,k}, v_i^{t+1};\xi_i) - \nabla_u F_i(u_i^{t,k}, v_i^{t+1})+ \nabla_u F_i(u_i^{t,k}, v_i^{t+1}) \\
        &  - \nabla_u F_i(u_i^{t}, V^{t+1}) +\nabla_u F_i(u_i^{t}, V^{t+1}) \Big) \Big\|^2\\
        & \leq \text{I} + \text{II}. \nonumber
    \end{split}
\end{equation}
Where 
\begin{equation}
    \text{I} = (1+ \frac{1}{2K_u -1}) \frac{1}{m}\sum_{i=1}^m \E \Big\| u_i^{t,k} - u_i^{t} \Big\|^2, \nonumber
\end{equation}
and 
\begin{equation}
    \text{II} =  \frac{2K_u^2\eta_u^2}{m}\sum_{i=1}^m \E \Big\| \nabla_u F_i(u_i^{t,k}, v_i^{t+1};\xi_i) - \nabla_u F_i(u_i^{t,k}, v_i^{t+1}) + \nabla_u F_i(u_i^{t,k}, v_i^{t+1}) - \nabla_u F(u_i^{t}, V^{t+1}) +\nabla_u F(u_i^{t}, V^{t+1}) \Big) \Big\|^2 . \nonumber
\end{equation}
% For I, we use assumption 1 and generate the following:
% \begin{align}
% \begin{split}
%     \text{I} & \leq (1+ \frac{1}{2K_u -1}) \frac{1}{m}\sum_{i=1}^m \E \Big\| u_i^{t,k} - u_i^{t}\Big\|^2  \\
%     & \leq (1+ \frac{1}{2K_u -1}) \frac{1}{m}\sum_{i=1}^m \E \Big\| u_i^{t,k} - u_i^{t}\Big\|^2 . \nonumber
% \end{split}
% \end{align}
For II, we use assumptions 2-3 and generate the following:
\begin{align}
\begin{split}
    \text{II} = 6\eta_u^2K_u \Big(\sigma_u^2 + \delta^2 + \frac{1}{m}\sum_{i=1}^m  \E \Big\|\nabla_u F(u_i^{t},V^{t+1}) \Big\|^2 \Big). \nonumber
\end{split}
\end{align}
Therefore, the recursion from $j=0$ to $K_u-1$ can generate:
\begin{equation}
    \begin{split}
        \frac{1}{m}\sum_{i=1}^m \E \bigl\| u_i^{t,k} - u_i^t\bigr \|^2 
        & \leq \sum_{j=0}^{K_u-1}(1+ \frac{1}{2K_u -1})^j \text{II} \\
        & \leq (2K_u-1)\Big[(1+ \frac{1}{2K_u -1})^K_u -1 \Big]\text{II}\\
        & \overset{a)}{\leq}3K_u \text{II} \\
        & \leq 18\eta_u^2K_u^2\Big(\sigma_u^2+\delta^2+\frac{1}{m}\sum_{i=1}^m\E \bigl\| \nabla_u F(u_i^{t},V^{t+1}) \bigr \|^2\Big), \nonumber
    \end{split}
\end{equation}
where a) uses $1+ \frac{1}{2K_u -1} \leq 2$ and $(1+ \frac{1}{2K_u  -1})^{2K_u  \cdot \frac{1}{2}} \leq \sqrt{5} < \frac{5}{2}$ for any $K_u \ge 1$.
\end{proof}
\end{lemma}
\begin{lemma}[Local update for shared model $u_i$ in DFedSalt]
\label{le:u_i_t_k}
Assume that assumptions 1-3 hold, for all clients $i \in \{1,2,...,m\}$ and local iteration steps $k \in \{0,1,..., K_u-1\}$, we can get
\begin{equation}
    \frac{1}{m}\sum_{i=1}^m \E \bigl\| u_i^{t,k} - u_i^t\bigr \|^2 \leq 6\eta_u^2K_uL^2_u\rho^2+18\eta_u^2K_u^2\Big(\sigma_u^2+\delta^2+\frac{1}{m}\sum_{i=1}^m\E \bigl\| \nabla_u F(u_i^t,V^{t+1}) \bigr \|^2\Big). \nonumber
\end{equation}
\begin{proof}
\begin{equation}
    \begin{split}
        \frac{1}{m}\sum_{i=1}^m \E \bigl\| u_i^{t,k+1} - u_i^t\bigr \|^2 
        & \leq \frac{1}{m}\sum_{i=1}^m  \E \Big\| u_i^{t,k} - \eta_u \nabla_u F_i(u_i^{t,k}+\epsilon(u_i^{t,k}), v_i^{t+1};\xi_i) - u_i^t \Big\|^2\\
        & \leq \frac{1}{m}\sum_{i=1}^m  \E \Big\| u_i^{t,k} - u_i^{t} - \eta_u \Big( \nabla_u F_i(u_i^{t,k}+\epsilon(u_i^{t,k}), v_i^{t+1};\xi_i) - \nabla_u F_i(u_i^{t,k}, v_i^{t+1};\xi_i) \\
        & + \nabla_u F_i(u_i^{t,k}, v_i^{t+1};\xi_i) - \nabla_u F_i(u_i^{t,k}, v_i^{t+1})+ \nabla_u F_i(u_i^{t,k}, v_i^{t+1}) \\
        &  - \nabla_u F_i(u_i^{t}, V^{t+1}) +\nabla_u F_i(u_i^{t}, V^{t+1}) \Big) \Big\|^2\\
        & \leq \text{I} + \text{II}. \nonumber
    \end{split}
\end{equation}
Where 
\begin{equation}
    \text{I} = (1+ \frac{1}{2K_u -1}) \frac{1}{m}\sum_{i=1}^m \E \Big\| u_i^{t,k} - u_i^{t} -  \eta_u \Big( \nabla_u F_i(u_i^{t,k}+\epsilon(u_i^{t,k}), v_i^{t+1};\xi_i) - \nabla_u F_i(u_i^{t,k}, v_i^{t+1};\xi_i)\Big)\Big\|^2, \nonumber
\end{equation}
and 
\begin{equation}
    \text{II} =  \frac{2K_u^2\eta_u^2}{m}\sum_{i=1}^m \E \Big\| \nabla_u F_i(u_i^{t,k}, v_i^{t+1};\xi_i) - \nabla_u F_i(u_i^{t,k}, v_i^{t+1}) + \nabla_u F_i(u_i^{t,k}, v_i^{t+1}) - \nabla_u F(u_i^{t}, V^{t+1}) +\nabla_u F(u_i^{t}, V^{t+1}) \Big) \Big\|^2 . \nonumber
\end{equation}
For I, we use assumption 1 and generate the following:
\begin{align}
\begin{split}
    \text{I} & \leq (1+ \frac{1}{2K_u -1}) \frac{1}{m}\sum_{i=1}^m \Big( \E \Big\| u_i^{t,k} - u_i^{t}\Big\|^2 + \eta_u^2L_u^2 \E \Big\| \rho \frac{\nabla_u F_i({u}_{i}^{t, k}, {v}_{i}^{t + 1};\xi_i)}{ \| \nabla_u F_i({u}_{i}^{t, k}, {v}_{i}^{t + 1};\xi_i)  \|_2}\Big\|^2\Big) \\
    & \leq (1+ \frac{1}{2K_u -1})\Big( \frac{1}{m}\sum_{i=1}^m \E \Big\| u_i^{t,k} - u_i^{t}\Big\|^2 + \eta_u^2L_u^2\rho^2\Big). \nonumber
\end{split}
\end{align}
For II, we use assumption 2-3 and generate the following:
\begin{align}
\begin{split}
    \text{II} = 6\eta_u^2K_u \Big(\sigma_u^2 + \delta^2 + \frac{1}{m}\sum_{i=1}^m  \E \Big\|\nabla_u F(u_i^{t},V^{t+1}) \Big\|^2 \Big). \nonumber
\end{split}
\end{align}
Therefore, the recursion from $j=0$ to $K_u-1$ can generate:
\begin{equation}
    \begin{split}
        \frac{1}{m}\sum_{i=1}^m \E \bigl\| u_i^{t,k} - u_i^t\bigr \|^2 
        & \leq \sum_{j=0}^{K_u-1}(1+ \frac{1}{2K_u -1})^j \Big[(1+ \frac{1}{2K_u -1})\eta_u^2L_u^2\rho^2 + \text{II} \Big]\\
        & \leq (2K_u-1)\Big[(1+ \frac{1}{2K_u -1})^K_u -1 \Big]\Big((1+ \frac{1}{2K_u -1})\eta_u^2L_u^2\rho^2 + \text{II} \Big)\\
        & \overset{a)}{\leq}3K_u (\text{II}  + 2\eta^2L_u^2\rho^2)\\
        & \leq 6\eta_u^2K_uL^2_u\rho^2+18\eta_u^2K_u^2\Big(\sigma_u^2+\delta^2+\frac{1}{m}\sum_{i=1}^m\E \bigl\| \nabla_u F(u_i^{t},V^{t+1}) \bigr \|^2\Big), \nonumber
    \end{split}
\end{equation}
where a) uses $1+ \frac{1}{2K_u -1} \leq 2$ and $(1+ \frac{1}{2K_u  -1})^{2K_u  \cdot \frac{1}{2}} \leq \sqrt{5} < \frac{5}{2}$ for any $K_u \ge 1$.
\end{proof}
\end{lemma}

\begin{lemma}[Shared model shift in DFedAlt]
\label{le:gossip_alt}
Assume that assumptions 1-3 hold, for all clients $i \in \{1,2,...,m\}$ and local iteration steps $k \in \{1,2,..., K_u\}$, we can get
\begin{equation}
    \frac{1}{m}\sum_{i=1}^m \E \|u_i^t - \Bar{u}^t\|^2 \leq \frac{18\eta_u^2K_u^2}{(1-\lambda)^2}\Big(\sigma_u^2+\delta^2 + \frac{1}{m}\sum_{i=1}^m \E \|\nabla_u F(u_i^t, V^{t+1})\|^2 \Big). \nonumber
\end{equation}
\begin{proof}
    Inspired by Lemma D.2 in \cite{shi2023improving} and Lemma 4 in \cite{sun2022decentralized}, according to Lemma \ref{mi}, we can generate 
    \begin{align*}
        \mathbb{E}\|{U}^{t}-{\bf P}{U}^{t}\|^2\leq &
        \E \|\sum_{j=0}^{t-1}({\bf P}-{\bf W}^{t-1-j}){\bf \zeta}^j\|^2\\
        &\leq \sum_{j=0}^{t-1}\|{\bf P}-{\bf W}^{t-1-j}\|_{\textrm{op}}\|{\bf \zeta}^j\|\\
        & \leq (\sum_{j=0}^{t-1}\lambda^{t-1-j}\|{\bf \zeta}^j\|)^2\\
        & \leq \mathbb{E}(\sum_{j=0}^{t-1}\lambda^{\frac{t-1-j}{2}}\cdot \lambda^{\frac{t-1-j}{2}}\|{\bf \zeta}^j\|)^2\\
        &\leq (\sum_{j=0}^{t-1}\lambda^{t-1-j})(\sum_{j=0}^{t-1} \lambda^{t-1-j}\mathbb{E}\|{\bf \zeta}^j\|^2),
\end{align*}
where $U^t = [u^t_1, u^t_2,..., u^t_m]^T \in \mathbb{R}^{m\times d}$ and $$\mathbb{E}\|{\bf \zeta}^j\|^2\leq \|{\bf W}\|^2\cdot\mathbb{E}\|{U}^{j}-Z^{j}\|^2\leq \mathbb{E}\|{U}^{j}-Z^{j}\|^2.$$
Note that $Z^t = [z^t_1, z^t_2,..., z^t_m]^T \in \mathbb{R}^{m\times d}$. According to Lemma \ref{le:u_i_t_k_alt}, for any $j$, we have
\begin{align*}
&\mathbb{E}\|U^{j}-Z^{j}\|^2 \leq 18\eta_u^2K_u^2m\Big(\sigma_u^2+\delta^2+\frac{1}{m}\sum_{i=1}^m\E \bigl\| \nabla_u F(u_i^t,V^{t+1}) \bigr \|^2\Big).
\end{align*}
After that, 
 \begin{align*}
        &\mathbb{E}\|{U}^{t}-{\bf P}{U}^{t}\|^2 \leq \frac{18\eta_u^2K_u^2}{(1-\lambda)^2}\Big(\sigma_u^2+\delta^2 + \frac{1}{m}\sum_{i=1}^m \E \|\nabla_u F(u_i^t, V^{t+1})\|^2 \Big).
\end{align*}
The fact is that $U^{t}-{\bf P}U^{t}=\left(
                                                \begin{array}{c}
                                                  u^t_1- \Bar{u^t}\\
                                                   u^t_2- \Bar{u^t} \\
                                                  \vdots \\
                                                   u^t_m- \Bar{u^t} \\
                                                \end{array}
                                              \right)
$ is the result needed to prove.
\end{proof}
\end{lemma}

\begin{lemma}[Shared model shift in DFedSalt]
\label{le:gossip}
Assume that assumptions 1-3 hold, for all clients $i \in \{1,2,...,m\}$ and local iteration steps $k \in \{1,2,..., K_u\}$, we can get
\begin{equation}
    \frac{1}{m}\sum_{i=1}^m \E \|u_i^t - \Bar{u}^t\|^2 \leq \frac{6\eta_u^2K_u}{(1-\lambda)^2}\Big[L_u^2\rho^2 + 3K_u\Big(\sigma_u^2+\delta^2 + \frac{1}{m}\sum_{i=1}^m \E \|\nabla_u F(u_i^t, V^{t+1})\|^2 \Big)\Big]. \nonumber
\end{equation}
\begin{proof}
   Based on Lemma \ref{le:gossip_alt} and according to Lemma \ref{le:u_i_t_k}, for any $j$, we have
\begin{align*}
&\mathbb{E}\|U^{j}-Z^{j}\|^2 \leq m\Big(6\eta_u^2K_uL^2_u\rho^2+18\eta_u^2K_u^2\Big(\sigma_u^2+\delta^2+\frac{1}{m}\sum_{i=1}^m\E \bigl\| \nabla_u F(u_i^t,V^{t+1}) \bigr \|^2\Big)\Big).
\end{align*}
After that, 
 \begin{align*}
        &\mathbb{E}\|{U}^{t}-{\bf P}{U}^{t}\|^2 \leq \frac{6\eta_u^2K_u}{(1-\lambda)^2}\Big[L_u^2\rho^2 + 3K_u\Big(\sigma_u^2+\delta^2 + \frac{1}{m}\sum_{i=1}^m \E \|\nabla_u F(u_i^t, V^{t+1})\|^2 \Big)\Big].
\end{align*}
The fact is that $U^{t}-{\bf P}U^{t}=\left(
                                                \begin{array}{c}
                                                  u^t_1- \Bar{u^t}\\
                                                   u^t_2- \Bar{u^t} \\
                                                  \vdots \\
                                                   u^t_m- \Bar{u^t} \\
                                                \end{array}
                                              \right)
$ is the result needed to prove.
\end{proof}
\end{lemma}

\subsection{Proof of Convergence Analysis}
\textbf{Proof Outline and the Challenge of Dependent Random Variables.}
    We start with 
    \begin{align} 
        \begin{aligned}
        F\left(\Bar{u}^{t+1}, V^{t+1}\right)
        - F\left(\Bar{u}^{t}, V^{t}\right)
        =&\, F\left(\Bar{u}^{t}, V^{t+1}\right)
        - F\left(\Bar{u}^{t}, V^{t}\right) \\
        &+ F\left(\Bar{u}^{t+1}, V^{t+1}\right)
        - F\left(\Bar{u}^{t}, V^{t+1}\right) \,.
        \end{aligned}
    \end{align}
    The first line corresponds to the effect of the $v$-step and the second line to the $u$-step. The former is 
    \begin{equation}
        \begin{split}
            F\left(\Bar{u}^{t}, V^{t+1}\right)
        - F\left(\Bar{u}^{t}, V^{t}\right) & = \frac{1}{m}\sum_{i=1}^m \E \Big[F_i(\Bar{u}^t, v_i^{t+1} ) - F_i(\Bar{u}^t, v_i^{t} )\Big]\\
        & \le 
        \frac{1}{m}\sum_{i=1}^m \E \Big[
        \Big <\nabla_v F_i\left(\Bar{u}
        ^{t}, v^{t}_i\right), v^{t+1}_i - v^{t}_i \Big>
        + \frac{L_v}{2}\|v^{t+1}_i - v^{t}_i \|^2 \Big] \,.
        \end{split}
    \end{equation}
    It is easy to handle with standard techniques that rely on the smoothness of $F\left(u^{t}, \cdot\right)$. 
    The latter is more challenging. 
    In particular, the smoothness bound for the $u$-step gives us
    \begin{align}
        F&\left(\Bar{u}^{t+1}, V^{t+1}\right)
        - F\left(\Bar{u}^{t}, V^{t+1}\right)
        \le 
        \Big <\nabla_u F\left(\Bar{u}
        ^{t}, V^{t+1}\right), \Bar{u}^{t+1} - \Bar{u}^{t} \Big>
        + \frac{L_u}{2}\|\Bar{u}^{t+1} - \Bar{u}^{t}\|^2 \,.
    \end{align}
    
\subsubsection{Proof of Convergence Analysis for DFedAlt}

\textbf{Analysis of the $u$-Step.}
    \begin{align} 
        \begin{aligned}
        & \E \Big [F\left(\Bar{u}^{t+1}, V^{t+1}\right)
        - F\left(\Bar{u}^{t}, V^{t+1}\right) \Big]
         \le 
        \Big <\nabla_u F\left(\Bar{u}
        ^{t}, V^{t+1}\right), \Bar{u}^{t+1} - \Bar{u}^{t} \Big>
        + \frac{L_u}{2}\E\|\Bar{u}^{t+1} - \Bar{u}^{t}\|^2\\
        & \leq \frac{-\eta_u}{m}\sum_{i=1}^m\E\Big <\nabla_u F\left(\Bar{u}
        ^{t}, V^{t+1}\right), \sum_{k=0}^{K_u-1}\nabla_u F\left(u_i^{t,k}, v_i^{t+1}; \xi_i\right)\Big> 
        + \frac{L_u}{2}\E\|\Bar{u}^{t+1} - \Bar{u}^{t}\|^2\\
        & \leq -\eta_uK_u \E [\Delta_{\Bar{u}}^t] + \frac{\eta_u}{m}\sum_{i=1}^m\sum_{k=0}^{K_u-1} \E \Big< \nabla_u F\left(\Bar{u}^{t}, V^{t+1}\right), \nabla F\left(\Bar{u}^t, v_i^{t+1}\right) - \nabla_u F\left(u_i^{t,k}, v_i^{t+1}; \xi_i\right) \Big> + \frac{L_u}{2}\E\|\Bar{u}^{t+1} - \Bar{u}^{t}\|^2 \\
        & \overset{a)}{\leq} \frac{-\eta_uK_u }{2} \E [\Delta_{\Bar{u}}^t] + \underbrace{\frac{\eta_uL_u^2}{2m}\sum_{i=1}^m\sum_{k=0}^{K_u-1} \E \| u_i^{t,k} - \Bar{u}^t \|^2}_{\mathcal{T}_{1, u}} + \underbrace{\frac{L_u}{2}\E\|\Bar{u}^{t+1} - \Bar{u}^{t}\|^2}_{\mathcal{T}_{2, u}}.
        \end{aligned}
    \end{align}
    Where a) uses $\E\left[ \nabla_u F(u_i^{t,k}, v_i^{t+1}; \xi_i) \right]= \nabla_u F\left(u_i^{t,k}, v_i^{t+1}\right)$ and $\left<x, y\right> \leq \frac{1}{2}\|x\|^2 + \frac{1}{2}\|y\|^2 $ for vectors $x, y$ followed by $L_u$-smoothness.\\
    For $\mathcal{T}_{1, u}$, we can use Lemma \ref{le:gossip_alt}. 
    \begin{equation}\label{T1_a}
    \begin{split}
        \mathcal{T}_{1, u} \leq \frac{9 \eta_u^3K_u^2L_u^2 }{(1-\lambda)^2}\Bigg[\sigma_u^2+\delta^2+\underbrace{\frac{1}{m}\sum_{i=1}^m\E \bigl\| \nabla_u F(u_i^t,V^{t+1}) \bigr \|^2}_{\mathcal{T}_{3, u}}\Bigg]
    \end{split}
    \end{equation}
    For $\mathcal{T}_{3, u}$,
    \begin{equation}\label{T3_a}
    \begin{split}
        \mathcal{T}_{3, u} & \leq \frac{1}{m}\sum_{i=1}^m\E \bigl\| \nabla_u F(u_i^t,V^{t+1}) - \nabla_u F(\Bar{u}^t,V^{t+1}) +  \nabla_u F(\Bar{u}^t,V^{t+1})\bigr \|^2 \\
        & \leq  \frac{2L_u^2}{m}\sum_{i=1}^m\E \|u_i^t - \Bar{u}^t \|^2+ \frac{2}{m}\sum_{i=1}^m\E \|\nabla_u F(\Bar{u}^t,V^{t+1})\|^2 \\
        & {\leq } \frac{2L_u^2}{m}\sum_{i=1}^m\E \|u_i^t - \Bar{u}^t \|^2 + 2 \E [\Delta_{\Bar{u}}^t] ,
    \end{split}
    \end{equation}
    After that, combining Eq. (\ref{T1_a}) and (\ref{T3_a}) and assuming local learning rate $\eta_u \ll  \frac{1-\lambda}{3\sqrt{2K_uL_u}}$,
we can generate
\begin{equation}\label{T_1_u_a}
    \begin{split}
        \mathcal{T}_{1, u} & \leq 
        \frac{9 \eta_u^3K_u^2L_u^2 }{(1-\lambda)^2}\Bigg[\sigma_u^2+\delta^2+2\E [\Delta_{\Bar{u}}^t]\Bigg].
    \end{split}
\end{equation}
    Meanwhile, for $\mathcal{T}_{2, u}$,
    \begin{align} 
        \begin{aligned}
         \mathcal{T}_{2, u} & \leq \frac{\eta_u^2L_u}{2m} \sum_{i=1}^m\sum_{k=0}^{K_u-1}\Big \|\nabla_u F\left(u_i^{t,k}, v_i^{t+1}; \xi_i\right) - 
         \nabla_u F\left(u_i^{t}, v_i^{t+1}\right) + \nabla_u F\left(u_i^{t}, v_i^{t+1}\right) \\
        & - \nabla_u F\left(u_i^{t}, V^{t+1}\right) + \nabla_u F\left(u_i^{t}, V^{t+1}\right) - \nabla_u F\left(\Bar{u}^{t} , V^{t+1}\right) + \nabla_u F\left(\Bar{u}^{t}, V^{t+1}\right)\Big \|^2 \\
        &  \leq 2\eta_u^2K_uL_u\Big(  \sigma_u^2+\delta^2+\frac{L_u^2}{m}\sum_{i=1}^m\E \|u_i^{t}-\Bar{u}^{t}\|^2 + \E[\Delta_{\Bar{u}}^t] \Big)\\
        &  \leq 2\eta_u^2K_uL_u\Big(  \sigma_u^2+\delta^2+ \E[\Delta_{\Bar{u}}^t] \Big)+\underbrace{\frac{2\eta_u^2K_uL_u^3}{m}\sum_{i=1}^m\E \|u_i^{t}-\Bar{u}^{t}\|^2}_{\mathcal{T}_{4, u}} 
        \end{aligned}
    \end{align}
For $\mathcal{T}_{4, u}$, we can use Lemma \ref{le:gossip_alt}. 

After that, 
\begin{equation}
    \begin{split}
        \E \Big [F\left(\Bar{u}^{t+1}, V^{t+1}\right)
        - F\left(\Bar{u}^{t}, V^{t+1}\right) \Big]
         & \le \frac{-\eta_uK_u}{2}\E[\Delta_{\Bar{u}}^t] + \mathcal{T}_{1, u} + \mathcal{T}_{2, u}\\
         & \le \Big( \frac{-\eta_uK_u}{2} + 2\eta_u^2K_uL_u + \frac{18\eta_u^2K_u^2(2+\eta_uL_u^2)}{(1-\lambda)^2}\Big)\E[\Delta_{\Bar{u}}^t]\\
         &  + 2\eta_u^2K_uL_u(\sigma_u^2+\delta^2) + \frac{9\eta_u^2K_u^2(2+\eta_uL_u^2)}{(1-\lambda)^2} \Big( \sigma_u^2+\delta^2 \Big)
         .
    \end{split}
\end{equation}

\textbf{Analysis of the $v$-Step.}
\begin{equation}
        \begin{split}
           \E \Big [ F\left(\Bar{u}^{t}, V^{t+1}\right)
        - F\left(\Bar{u}^{t}, V^{t}\right) \Big]
        &  \le  \underbrace{\frac{1}{m}\sum_{i=1}^m \E 
        \Big <\nabla_v F_i\left(\Bar{u}
        ^{t}, v^{t}_i\right), v^{t+1}_i - v^{t}_i \Big>}_{\mathcal{T}_{1, v}}
        + \underbrace{\frac{L_v}{2m}\sum_{i=1}^m \E  \|v^{t+1}_i - v^{t}_i \|^2 }_{\mathcal{T}_{2, v}}.
        \end{split}
    \end{equation}
For $\mathcal{T}_{1, v}$, 
\begin{equation}\label{eq:T_1_a}
    \begin{split}
       \mathcal{T}_{1, v} & \leq \frac{1}{m}\sum_{i=1}^m \E 
        \Big <\nabla_v F_i\left(\Bar{u}
        ^{t}, v^{t}_i\right) - \nabla_v F_i\left(u_i
        ^{t}, v^{t}_i\right) + \nabla_v F_i\left(u_i
        ^{t}, v^{t}_i\right), -\eta_v \sum_{k=0}^{K_v-1} \E \nabla_v F_i(u_i^t, v^{t}_i; \xi_i) \Big> \\
        & \overset{a)}{\leq} \frac{-\eta_vK_v}{m}\sum_{i=1}^m \E \| \nabla_v F_i(u_i^t, v^{t}_i) \|^2 + \frac{1}{m}\sum_{i=1}^m \E 
        \Big <\nabla_v F_i\left(\Bar{u}
        ^{t}, v^{t}_i\right) - \nabla_v F_i\left(u_i
        ^{t}, v^{t}_i\right), v^{t+1}_i - v^{t}_i \Big> \\
        & \overset{b)}{\leq} -\eta_vK_v \E [\Delta_v^t] + \underbrace{\frac{L_{vu}^2}{2m}\sum_{i=1}^m \E \|\Bar{u}^t-u_i^{t}\|^2}_{\mathcal{T}_{3, v}} + \underbrace{\frac{1}{2m}\sum_{i=1}^m \E \|v^{t+1}_i - v^{t}_i\|^2}_{\frac{1}{L_v}\mathcal{T}_{2, v}},
    \end{split}
\end{equation}
where a) and b) is get from the unbiased expectation property of $\nabla_v F_i(u_i^{t},v^{t}_i; \xi_i)$  and $<x, y> \leq \frac{1}{2}(\|x\|^2+\|y\|^2)$, respectively.

For $\mathcal{T}_{2, v}$, according to Lemma \ref{le:v}, we have 
\begin{equation}\label{eq:T_2_a}
    \begin{split}
       \mathcal{T}_{2, v} & \leq \frac{L_v}{2}\Big( \frac{16 \eta_v^2 K_v^2}{m}\sum_{i=1}^m \E \|\nabla_v F_i(u_i^t, v^{t}_i)\|^2 + 8 \eta_v^2 K_v^2 \sigma_v^2\Big)\\
       & \leq   8L_v\eta_v^2 K_v^2 \E [\Delta_v^t] + 4L_v\eta_v^2 K_v^2 \sigma_v^2.
    \end{split}
\end{equation}
For $\mathcal{T}_{3, v}$, according to Eq. (\ref{T_1_u_a}), we have 
\begin{equation}\label{eq:T_3_a}
    \frac{L_{vu}^2}{2m}\sum_{i=1}^m \E \|\Bar{u}^t-u_i^{t}\|^2 \leq \frac{L_{vu}^2 }{(1-\lambda)^2}\Bigg[18\eta_u^2K_u^2\Big(\sigma_u^2+\delta^2+2\E [\Delta_{\Bar{u}}^t]\Big)\Bigg].
\end{equation}
After that, summing Eq. (\ref{eq:T_1_a}), (\ref{eq:T_2_a}), and (\ref{eq:T_3_a}), we have
\begin{equation}
    \begin{split}
      \E \Big [ F\left(\Bar{u}^{t}, V^{t+1}\right)
        - F\left(\Bar{u}^{t}, V^{t}\right) \Big]  
        &  \le \Big(-\eta_vK_v + 8\eta_v^2K_v^2(L_v+1)\Big)\E [\Delta_v^t] +  4\eta_v^2K_v^2\sigma_v^2(L_v+1) \\
        & + \frac{L_{vu}^2 }{(1-\lambda)^2}\Bigg[18\eta_u^2K_u^2\Big(\sigma_u^2+\delta^2+2\E [\Delta_{\Bar{u}}^t]\Big)\Bigg].
    \end{split}
\end{equation}
% Using Eq. (\ref{T_1_u}), we have
\textbf{Obtaining the Final Convergence Bound.} 
\begin{align} \label{eq:pfl-am:pf:2}
        \begin{aligned}
       \E \Big [   F\left(\Bar{u}^{t+1}, V^{t+1}\right)
        - F\left(\Bar{u}^{t}, V^{t}\right)  \Big]  
        =&\,\E \Big [  F\left(\Bar{u}^{t}, V^{t+1}\right)
        - F\left(\Bar{u}^{t}, V^{t}\right) + F\left(\Bar{u}^{t+1}, V^{t+1}\right)
        - F\left(\Bar{u}^{t}, V^{t+1}\right)  \Big]   \\
        & \leq \Big( \frac{-\eta_uK_u}{2} + 2\eta_u^2K_uL_u + \frac{18\eta_u^2K_u^2(2+\eta_uL_u^2)}{(1-\lambda)^2}\Big)\E[\Delta_{\Bar{u}}^t]\\
         &  + 2\eta_u^2K_uL_u(\sigma_u^2+\delta^2) + \frac{9\eta_u^2K_u^2(2+\eta_uL_u^2)}{(1-\lambda)^2} \Big( \sigma_u^2+\delta^2 \Big) \\
         & + \Big(-\eta_vK_v + 8\eta_v^2K_v^2(L_v+1)\Big)\E [\Delta_v^t] +  4\eta_v^2K_v^2\sigma_v^2(L_v+1) \\
        & + \frac{ 18\eta_u^2L_{vu}^2K_u^2\Big(\sigma_u^2+\delta^2+2\E [\Delta_{\Bar{u}}^t]\Big)}{(1-\lambda)^2}.
        \end{aligned}
    \end{align}
Summing from $t=1$ to $T$, assume the local learning rates satisfy $\eta_u=\mathcal{O}({1}/{L_uK_u\sqrt{T}}), \eta_v=\mathcal{O}({1}/{L_vK_v\sqrt{T}})$, $F^{*}$ is denoted as the minimal value of $F$, i.e., $F(\bar{u}, V)\ge F^*$ for all $\bar{u} \in \mathbb{R}^{d}$, and $V=(v_1,\ldots,v_m)\in\mathbb{R}^{d_1+\ldots+d_m}$. We can generate
\begin{equation}
    \begin{split}
        \frac{1}{T}\sum_{i=1}^T \bigl(\frac{1}{L_u} \E \bigl[\Delta_{\bar{u}}^t \bigr] + \frac{1}{L_v} \E [\Delta_{v}^t \bigr] \bigr) 
        & \leq \mathcal{O}\Big(\frac{F(\bar{u}^1, V^1) - F^*}{\sqrt{T}} + \frac{\sigma_v^2(L_v+1)}{L_v^2\sqrt{T}}\\
        & + \frac{L_{vu}^2(\sigma_u^2+\delta^2)}{L_u^2\sqrt{T}} +  \frac{\sigma_u^2+\delta^2}{L_uT(1-\lambda)^2} \Big) .
    \end{split}
\end{equation}
Assume that
\begin{align}
        \sigma_1^2 = \frac{\sigma_v^2(L_v+1)}{L_v^2} + \frac{L_{vu}^2(\sigma_u^2+\delta^2)}{L_u^2} = \frac{\sigma_v^2(L_v+1)}{L_v^2} + \frac{\chi^2L_v(\sigma_u^2+\delta^2)}{L_u} \, ,~~~
        \sigma_2^2 = \frac{\sigma_u^2+\delta^2}{L_u}\, .\nonumber
\end{align}
Then, we have the final convergence bound:
\begin{equation} 
\small
\frac{1}{T}\sum_{i=1}^T \bigl(\frac{1}{L_u} \E \bigl[\Delta_{\bar{u}}^t \bigr] + \frac{1}{L_v} \E [\Delta_{v}^t \bigr] \bigr) \leq \mathcal{O}\Big(\frac{F(\bar{u}^1, V^1) - F^*}{\sqrt{T}} + \frac{\sigma_1^2}{\sqrt{T}} + \frac{\sigma_2^2}{T(1-\lambda)^2} \Big).
\end{equation}

\subsubsection{Proof of Convergence Analysis for DFedSalt}

\textbf{Analysis of the $u$-Step.}
 \begin{align} 
        \begin{aligned}
        & \E \Big [F\left(\Bar{u}^{t+1}, V^{t+1}\right)
        - F\left(\Bar{u}^{t}, V^{t+1}\right) \Big]
         \le 
        \Big <\nabla_u F\left(\Bar{u}
        ^{t}, V^{t+1}\right), \Bar{u}^{t+1} - \Bar{u}^{t} \Big>
        + \frac{L_u}{2}\E\|\Bar{u}^{t+1} - \Bar{u}^{t}\|^2\\
        & \leq \frac{-\eta_u}{m}\sum_{i=1}^m\E\Big <\nabla_u F\left(\Bar{u}
        ^{t}, V^{t+1}\right), \sum_{k=0}^{K_u-1}\nabla_u F\left(u_i^{t,k}, v_i^{t+1}; \xi_i\right)\Big> 
        + \frac{L_u}{2}\E\|\Bar{u}^{t+1} - \Bar{u}^{t}\|^2\\
        & \leq -\eta_uK_u \E [\Delta_{\Bar{u}}^t] + \frac{\eta_u}{m}\sum_{i=1}^m\sum_{k=0}^{K_u-1} \E \Big< \nabla_u F\left(\Bar{u}^{t}, V^{t+1}\right), \nabla F\left(\Bar{u}^t, v_i^{t+1}\right) - \nabla_u F\left(u_i^{t,k}, v_i^{t+1}; \xi_i\right) \Big> + \frac{L_u}{2}\E\|\Bar{u}^{t+1} - \Bar{u}^{t}\|^2 \\
        & \overset{a)}{\leq} \frac{-\eta_uK_u }{2} \E [\Delta_{\Bar{u}}^t] + \underbrace{\frac{\eta_uL_u^2}{2m}\sum_{i=1}^m\sum_{k=0}^{K_u-1} \E \| u_i^{t,k} - \Bar{u}^t \|^2}_{\mathcal{T}_{1, u}} + \underbrace{\frac{L_u}{2}\E\|\Bar{u}^{t+1} - \Bar{u}^{t}\|^2}_{\mathcal{T}_{2, u}}.
        \end{aligned}
    \end{align}
    Where a) uses $\E\left[ \nabla_u F(u_i^{t,k}, v_i^{t+1}; \xi_i) \right]= \nabla_u F\left(u_i^{t,k}, v_i^{t+1}\right)$ and $\left<x, y\right> \leq \frac{1}{2}\|x\|^2 + \frac{1}{2}\|y\|^2 $ for vectors $x, y$ followed by $L_u$-smoothness.\\
    For $\mathcal{T}_{1, u}$, we can use Lemma \ref{le:gossip}. 
    \begin{equation}\label{T1}
    \begin{split}
        \mathcal{T}_{1, u} \leq \frac{3 \eta_u^3K_uL_u^2 }{(1-\lambda)^2}\Bigg[
        L^2_u\rho^2+3K_u\Big(\sigma_u^2+\delta^2+\underbrace{\frac{1}{m}\sum_{i=1}^m\E \bigl\| \nabla_u F(u_i^t,V^{t+1}) \bigr \|^2}_{\mathcal{T}_{3, u}}\Big)\Bigg]
    \end{split}
    \end{equation}
    For $\mathcal{T}_{3, u}$,
    \begin{equation}\label{T3}
    \begin{split}
        \mathcal{T}_{3, u} & \leq \frac{1}{m}\sum_{i=1}^m\E \bigl\| \nabla_u F(u_i^t,V^{t+1}) - \nabla_u F(\Bar{u}^t,V^{t+1}) +  \nabla_u F(\Bar{u}^t,V^{t+1})\bigr \|^2 \\
        & \leq  \frac{2L_u^2}{m}\sum_{i=1}^m\E \|u_i^t - \Bar{u}^t \|^2+ \frac{2}{m}\sum_{i=1}^m\E \|\nabla_u F(\Bar{u}^t,V^{t+1})\|^2 \\
        & {\leq } \frac{2L_u^2}{m}\sum_{i=1}^m\E \|u_i^t - \Bar{u}^t \|^2 + 2 \E [\Delta_{\Bar{u}}^t] ,
    \end{split}
    \end{equation}
    After that, combining Eq. (\ref{T1}) and (\ref{T3}) and assuming local learning rate $\eta_u \ll  \frac{1-\lambda}{3\sqrt{2K_uL_u}}$,
we can generate
\begin{equation}\label{T_1_u}
    \begin{split}
        \mathcal{T}_{1, u} & \leq 
         \frac{3 \eta_u^3K_uL_u^2 }{(1-\lambda)^2}\Bigg[
        L^2_u\rho^2+3K_u\Big(\sigma_u^2+\delta^2+2\E [\Delta_{\Bar{u}}^t]\Big)\Bigg].
    \end{split}
\end{equation}
    Meanwhile, for $\mathcal{T}_{2, u}$,
    \begin{align} 
        \begin{aligned}
         \mathcal{T}_{2, u} & \leq \frac{\eta_u^2L_u}{2m} \sum_{i=1}^m\sum_{k=0}^{K_u-1}\Big \|\nabla_u F\left(u_i^{t,k}+\epsilon(u_i^{t,k}), v_i^{t+1}; \xi_i\right) - \nabla_u F\left(u_i^{t,k}, v_i^{t+1}; \xi_i\right) \\
        & + \nabla_u F\left(u_i^{t,k}, v_i^{t+1}; \xi_i\right) - 
         \nabla_u F\left(u_i^{t}, v_i^{t+1}\right) + \nabla_u F\left(u_i^{t}, v_i^{t+1}\right) - \nabla_u F\left(u_i^{t}, V^{t+1}\right) + \nabla_u F\left(u_i^{t}, V^{t+1}\right) \\
        & - \nabla_u F\left(\Bar{u}^{t} , V^{t+1}\right) + \nabla_u F\left(\Bar{u}^{t}, V^{t+1}\right)\Big \|^2 \\
        &  \leq \frac{5}{2}\eta_u^2K_uL_u\Big( L_u^2\rho^2+ \sigma_u^2+\delta^2+\frac{L_u^2}{m}\sum_{i=1}^m\E \|u_i^{t}-\Bar{u}^{t}\|^2 + \E[\Delta_{\Bar{u}}^t] \Big)\\
        &  \leq \frac{5}{2}\eta_u^2K_uL_u\Big( L_u^2\rho^2+ \sigma_u^2+\delta^2+ \E[\Delta_{\Bar{u}}^t] \Big)+\underbrace{\frac{5\eta_u^2K_uL_u^3}{2m}\sum_{i=1}^m\E \|u_i^{t}-\Bar{u}^{t}\|^2}_{\mathcal{T}_{4, u}} 
        \end{aligned}
    \end{align}
For $\mathcal{T}_{4, u}$, we can use Lemma \ref{le:gossip}. 

After that, 
\begin{equation}
    \begin{split}
        \E \Big [F\left(\Bar{u}^{t+1}, V^{t+1}\right)
        - F\left(\Bar{u}^{t}, V^{t+1}\right) \Big]
         & \le \frac{-\eta_uK_u}{2}\E[\Delta_{\Bar{u}}^t] + \mathcal{T}_{1, u} + \mathcal{T}_{2, u}\\
         & \le \Big( \frac{-\eta_uK_u}{2} + \frac{5}{2}\eta_u^2K_uL_u + \frac{18\eta_u^3K_u^2L_u^2(1+5\eta_uK_uL_u)}{(1-\lambda)^2}\Big)\E[\Delta_{\Bar{u}}^t]\\
         &  + \frac{3 \eta_u^3K_uL_u^2 }{(1-\lambda)^2}\Bigg[
        L^2_u\rho^2+3K_u\Big(\sigma_u^2+\delta^2\Big)\Bigg] + \frac{5}{2}\eta_u^2K_uL_u\Big( L_u^2\rho^2+ \sigma_u^2+\delta^2 \Big)\\
        & + \frac{15 \eta_u^4K_u^2L_u^3 }{(1-\lambda)^2}\Bigg[
        L^2_u\rho^2+3K_u\Big(\sigma_u^2+\delta^2\Big)\Bigg] 
         .
    \end{split}
\end{equation}

\textbf{Analysis of the $v$-Step.}
\begin{equation}
        \begin{split}
           \E \Big [ F\left(\Bar{u}^{t}, V^{t+1}\right)
        - F\left(\Bar{u}^{t}, V^{t}\right) \Big]
        &  \le  \underbrace{\frac{1}{m}\sum_{i=1}^m \E 
        \Big <\nabla_v F_i\left(\Bar{u}
        ^{t}, v^{t}_i\right), v^{t+1}_i - v^{t}_i \Big>}_{\mathcal{T}_{1, v}}
        + \underbrace{\frac{L_v}{2m}\sum_{i=1}^m \E  \|v^{t+1}_i - v^{t}_i \|^2 }_{\mathcal{T}_{2, v}}.
        \end{split}
    \end{equation}
For $\mathcal{T}_{1, v}$, 
\begin{equation}\label{eq:T_1}
    \begin{split}
       \mathcal{T}_{1, v} & \leq \frac{1}{m}\sum_{i=1}^m \E 
        \Big <\nabla_v F_i\left(\Bar{u}
        ^{t}, v^{t}_i\right) - \nabla_v F_i\left(u_i
        ^{t}, v^{t}_i\right) + \nabla_v F_i\left(u_i
        ^{t}, v^{t}_i\right), -\eta_v \sum_{k=0}^{K_v-1} \E \nabla_v F_i(u_i^t, v^{t}_i; \xi_i) \Big> \\
        & \overset{a)}{\leq} \frac{-\eta_vK_v}{m}\sum_{i=1}^m \E \| \nabla_v F_i(u_i^t, v^{t}_i) \|^2 + \frac{1}{m}\sum_{i=1}^m \E 
        \Big <\nabla_v F_i\left(\Bar{u}
        ^{t}, v^{t}_i\right) - \nabla_v F_i\left(u_i
        ^{t}, v^{t}_i\right), v^{t+1}_i - v^{t}_i \Big> \\
        & \overset{b)}{\leq} -\eta_vK_v \E [\Delta_v^t] + \underbrace{\frac{L_{vu}^2}{2m}\sum_{i=1}^m \E \|\Bar{u}^t-u_i^{t}\|^2}_{\mathcal{T}_{3, v}} + \underbrace{\frac{1}{2m}\sum_{i=1}^m \E \|v^{t+1}_i - v^{t}_i\|^2}_{\frac{1}{L_v}\mathcal{T}_{2, v}},
    \end{split}
\end{equation}
where a) and b) is get from the unbiased expectation property of $\nabla_v F_i(u_i^{t},v^{t}_i; \xi_i)$  and $<x, y> \leq \frac{1}{2}(\|x\|^2+\|y\|^2)$, respectively.

For $\mathcal{T}_{2, v}$, according to Lemma \ref{le:v}, we have 
\begin{equation}\label{eq:T_2}
    \begin{split}
       \mathcal{T}_{2, v} & \leq \frac{L_v}{2}\Big( \frac{16 \eta_v^2 K_v^2}{m}\sum_{i=1}^m \E \|\nabla_v F_i(u_i^t, v^{t}_i)\|^2 + 8 \eta_v^2 K_v^2 \sigma_v^2\Big)\\
       & \leq   8L_v\eta_v^2 K_v^2 \E [\Delta_v^t] + 4L_v\eta_v^2 K_v^2 \sigma_v^2.
    \end{split}
\end{equation}
For $\mathcal{T}_{3, v}$, according to Eq. (\ref{T_1_u}), we have 
\begin{equation}\label{eq:T_3}
    \frac{L_{vu}^2}{2m}\sum_{i=1}^m \E \|\Bar{u}^t-u_i^{t}\|^2 \leq \frac{L_{vu}^2 }{(1-\lambda)^2}\Bigg[3\eta_u^2K_uL^2_u\rho^2+9\eta_u^2K_u^2\Big(\sigma_u^2+\delta^2+2\E [\Delta_{\Bar{u}}^t]\Big)\Bigg].
\end{equation}
After that, summing Eq. (\ref{eq:T_1}), (\ref{eq:T_2}), and (\ref{eq:T_3}), we have
\begin{equation}
    \begin{split}
        \E \Big [ F\left(\Bar{u}^{t}, V^{t+1}\right)
        - F\left(\Bar{u}^{t}, V^{t}\right) \Big]
        &  \le \Big(-\eta_vK_v + 8\eta_v^2K_v^2(L_v+1)\Big)\E [\Delta_v^t] +  4\eta_v^2K_v^2\sigma_v^2(L_v+1) \\
        & + \frac{L_{vu}^2 }{(1-\lambda)^2}\Bigg[3\eta_u^2K_uL^2_u\rho^2+9\eta_u^2K_u^2\Big(\sigma_u^2+\delta^2+2\E [\Delta_{\Bar{u}}^t]\Big)\Bigg].
    \end{split}
\end{equation}
% Using Eq. (\ref{T_1_u}), we have
\textbf{Obtaining the Final Convergence Bound.} 
\begin{align} \label{eq:pfl-am:pf:1}
        \begin{aligned}
        \E \Big [   F\left(\Bar{u}^{t+1}, V^{t+1}\right)
        - F\left(\Bar{u}^{t}, V^{t}\right)  \Big]  
        =&\,\E \Big [ F\left(\Bar{u}^{t}, V^{t+1}\right)
        - F\left(\Bar{u}^{t}, V^{t}\right) + F\left(\Bar{u}^{t+1}, V^{t+1}\right)
        - F\left(\Bar{u}^{t}, V^{t+1}\right)\Big]   \\
        & \leq (\frac{\eta_u}{2}-\eta_uK_u + \frac{54\eta_u^3K_u^3L_u^2 }{(1-\lambda)^2})\E[\Delta_{\Bar{u}}^t] + \frac{5}{2}\eta_u^2K_uL_u\Big(L_u^2\rho^2   + \sigma_u^2+\delta^2\Big)\\
        & + \frac{3\eta_uK_uL_u^2 }{2(1-\lambda)^2}\Big(6\eta_u^2K_uL^2_u\rho^2+18\eta_u^2K_u^2(\sigma_u^2+\delta^2)\Big)\\
        & + \Big(-\eta_vK_v + 8\eta_v^2K_v^2(L_v+1)\Big)\E [\Delta_v^t] +  4\eta_v^2K_v^2\sigma_v^2(L_v+1) \\
        & + \frac{L_{vu}^2 }{(1-\lambda)^2}\Bigg[3\eta_u^2K_uL^2_u\rho^2+9\eta_u^2K_u^2\Big(\sigma_u^2+\delta^2+2\E [\Delta_{\Bar{u}}^t]\Big)\Bigg].
        \end{aligned}
    \end{align}
Summing from $t=1$ to $T$, assume the local learning rates satisfy $\eta_u=\mathcal{O}({1}/{L_uK_u\sqrt{T}}), \eta_v=\mathcal{O}({1}/{L_vK_v\sqrt{T}})$, $F^{*}$ is denoted as the minimal value of $F$, i.e., $F(\bar{u}, V)\ge F^*$ for all $\bar{u} \in \mathbb{R}^{d}$, and $V=(v_1,\ldots,v_m)\in\mathbb{R}^{d_1+\ldots+d_m}$. We can generate
\begin{equation}
    \begin{split}
        \frac{1}{T}\sum_{i=1}^T \bigl(\frac{1}{L_u} \E \bigl[\Delta_{\bar{u}}^t \bigr] + \frac{1}{L_v} \E [\Delta_{v}^t \bigr] \bigr) 
        & \leq \mathcal{O}\Big(\frac{F(\bar{u}^1, V^1) - F^*}{\sqrt{T}} 
        + \frac{\sigma_v^2(L_v+1)}{L_v^2\sqrt{T}} + \! \frac{L_u}{T} \\
        & + \frac{L^2_{vu}}{T^{1/2}(1\!-\!\lambda)^2}\Big( \frac{\rho^2}{K_u} + \frac{\sigma_u^2+\delta^2}{L_u^2} \Big)  + \frac{L_u}{T(1\!-\!\lambda)^2}\Big(\frac{\rho^2}{K_u} + \frac{\sigma_u^2+\delta^2}{L_u^2}\Big) \Big).
    \end{split}
\end{equation}
Assume that
\begin{align}
        \sigma^2 = \frac{\rho}{K_u} + \frac{\sigma_u^2+\delta^2}{L_u^2} \, .\nonumber
\end{align}
Then, we have the final convergence bound:
\begin{equation} 
\frac{1}{T}\sum_{i=1}^T \bigl(\frac{1}{L_u} \E \bigl[\Delta_{\bar{u}}^t \bigr] \!+\! \frac{1}{L_v} \E [\Delta_{v}^t \bigr] \bigr) \!\leq\! \mathcal{O}\Big(\frac{F(\bar{u}^1, V^1) \!-\! F^*}{\sqrt{T}} \!+ \frac{\sigma_v^2(L_v+1)}{L_v^2\sqrt{T}} + \! \frac{L_u}{T} + \frac{\sigma^2L^2_{vu}}{T^{1/2}(1\!-\!\lambda)^2}  + \frac{\sigma^2L_u}{K_u^2T(1\!-\!\lambda)^2} \Big).
\end{equation}
Furthermore,      
When the perturbation amplitude $\rho$ is
proportional to the learning rate, e.g., $\rho = \mathcal{O}(1/\sqrt{T})$,
the sequence of outputs $\Delta_{\bar{u}}^t $ and $\Delta_{v}^t$ generated by Alg. \ref{DFedSAlt}, we have:
\begin{align}
        \mathcal{O}\Big(\sigma^2 \Big) = \mathcal{O}\Big( \frac{\rho^2}{K_u} + \frac{\sigma_u^2+\delta^2}{L_u^2}\Big) =  \mathcal{O}\Big(\frac{1}{K_uT} + \frac{\sigma_u^2+\delta^2}{L_u^2} \Big)\, .
    \end{align}

\end{document}